\documentclass{article}

\usepackage[main,final]{neurips_2026}

\usepackage[utf8]{inputenc} %
\usepackage[T1]{fontenc}    %
\usepackage{hyperref}       %
\usepackage{url}            %
\usepackage{booktabs}       %
\usepackage{amsfonts}       %
\usepackage{nicefrac}       %
\usepackage{microtype}      %
\usepackage[table]{xcolor}         %

\usepackage{amsthm}
\newtheorem{theorem}{Theorem}
\newtheorem{lemma}{Lemma}
\newtheorem{corollary}{Corollary}

\usepackage{amsmath,amssymb,amsfonts}
\usepackage{booktabs}
\usepackage{tikz}

\usepackage{wrapfig} 

\usepackage{algorithm}
\usepackage{algorithmic}
\usepackage{graphicx}
\usepackage{tikz}
\usetikzlibrary{arrows.meta,positioning,calc}

\usepackage{tcolorbox}
\usepackage{listings}
\usepackage{xcolor}  
\usepackage{booktabs}
\usepackage{multirow}
\usepackage{makecell}
\usepackage{xcolor}
\usepackage{graphicx}
\usepackage[table]{xcolor}

\definecolor{ablationBlue}{RGB}{35, 92, 175}
\newcommand{\best}[1]{\textbf{#1}}
\newcommand{\second}[1]{\underline{#1}}

\definecolor{oursblue}{RGB}{230,242,252}
\definecolor{groupgray}{RGB}{238,238,238}
\newcommand{\grouprow}[1]{%
\rowcolor{groupgray}
\multicolumn{8}{l}{\textbf{#1}}\\
}

\newcommand{\grouprowmem}[1]{%
\rowcolor{groupgray}
\multicolumn{9}{l}{\textbf{#1}}\\
}

\definecolor{inactivegray}{RGB}{120,120,120}
\newcommand{\inactive}[1]{\textcolor{inactivegray}{#1}}

\newcommand{\abldrop}[1]{\textcolor{ablationBlue}{\tiny\,{(#1)}}}

\usepackage{tabularx}
\usepackage{array}
\newcolumntype{Y}{>{\raggedright\arraybackslash}X}

\usepackage{tabularx}
\usepackage{array}
\usepackage[table]{xcolor}

\definecolor{appBlue}{RGB}{34, 88, 153}
\definecolor{appDark}{RGB}{45, 50, 62}
\definecolor{appGray}{RGB}{92, 99, 112}
\definecolor{appLight}{RGB}{246, 249, 254}
\definecolor{appLine}{RGB}{216, 224, 236}

\newcolumntype{A}{>{\raggedright\arraybackslash}p{0.30\linewidth}}
\newcolumntype{B}{>{\raggedright\arraybackslash}X}

\newcommand{\appSection}[2]{%
\hyperref[#1]{%
\textbf{\textcolor{appBlue}{Appendix~\ref*{#1}}}}%
\par
\vspace{1pt}
\hyperref[#1]{%
\textbf{\textcolor{appDark}{#2}}}%
}

\newcommand{\appSub}[2]{%
\hyperref[#1]{%
\textcolor{appDark}{\ref*{#1}\quad #2}}%
\par
\vspace{2pt}
}

\title{When Should Agents Remember? Falsification-Gated Self-Evolution for LLM Agents}

\author{%
  \textbf{Runxuan Tang}$^{1,*}$ \quad \textbf{Haoyu Gao}$^{2,*}$ \quad \textbf{Yuyan Ding}$^{3}$ \quad \textbf{Junyi Yao}$^{4}$ \\
  \textbf{Zihao Zheng}$^{4}$ \quad \textbf{Zecheng Sheng}$^{5}$ \quad \textbf{Liwei Hou}$^{6,\dagger}$ \\
  \normalfont
  $^{1}$Nanyang Technological University \quad $^{2}$Georgia Institute of Technology \\
  $^{3}$University of Washington \quad $^{4}$Washington University in St. Louis \\
  $^{5}$The University of Sydney \quad $^{6}$National University of Singapore \\
  \texttt{TANG0609@e.ntu.edu.sg} \quad \texttt{gao.howard517@gmail.com} \\
  \texttt{yuyand2@uw.edu} \quad \texttt{j.yao@wustl.edu} \quad \texttt{z.zihaogary@wustl.edu} \\
  \texttt{zshe0457@uni.sydney.edu.au} \quad \texttt{liweihou@u.nus.edu} \\
  $^{*}$Equal contribution. \quad $^{\dagger}$Corresponding author.
}

\hypersetup{
  hidelinks,
  pdftitle={When Should Agents Remember? Falsification-Gated Self-Evolution for LLM Agents},
  pdfauthor={Runxuan Tang, Haoyu Gao, Yuyan Ding, Junyi Yao, Zihao Zheng, Zecheng Sheng, Liwei Hou}
}

\begin{document}

\maketitle

\begin{abstract}
LLM agents increasingly operate in long-horizon, verifier-rich environments, but current self-improvement mechanisms often write episode-derived lessons directly into memory, making local success an unreliable signal for reusable and safe knowledge.
This paper addresses the core question of when an experience-derived behavioral update should be admitted into persistent agent state.
We propose \textbf{Falsification-Gated Self-Evolution} (FGSE), a verifier-grounded framework that represents each candidate lesson as a structured hypothesis with an explicit precondition, behavior change, expected effect, and verifier.
FGSE applies the hypothesis only in a temporary state, tests it on target-transfer, falsification, and archived-regression probes, and then commits, refines, or rejects it according to measured gain and risk.
Across web, app, tool-use, and long-term memory benchmarks, FGSE achieves strong task performance, including 39.8 WebArena success rate, 78.8 AppWorld average completion, and 0.724/0.487 Pass$^1$ on $\tau$-Retail/$\tau$-Airline, while reducing falsification failures to 5.0\%, regression damage to 1.3\%, and harmful committed updates to 3.6\%.
These results suggest that testing memory updates as hypotheses supports more reliable long-term adaptation.
\end{abstract}

\section{Introduction}
\label{sec:introduction}

Large language model (LLM) agents are increasingly expected to operate in long-horizon, interactive environments where they must reason, call tools, update internal context, and recover from mistakes~\citep{yao2023react}. 
This capability supports web navigation, app-level tool use, and user-facing assistance~\citep{zhou2024webarena,trivedi2024appworld,yao2024taubench}; embodied interaction additionally requires grounding language in visual and gestural context~\citep{li2023touchline}. 
Continual deployment requires agents to solve current tasks and safely learn reusable behavioral knowledge from experience.

Recent work has made substantial progress toward self-improving agents through self-feedback and reflection~\citep{madaan2023selfrefine}, verbal reinforcement from past failures~\citep{shinn2023reflexion}, reusable skill memory~\citep{wang2023voyager,li2026unimpa}, agentic long-term memory organization~\citep{xu2025amem}, and evolving symbolic contexts or playbooks~\citep{zhou2025symboliclearning,zhang2026ace}.
These methods show that storing lessons, memories, playbooks, or revised contexts can improve future performance without necessarily updating model weights. 
An episode-derived update may improve tasks that resemble the triggering episode while overgeneralizing to superficially similar but semantically different situations, conflicting with previously successful behaviors, or accumulating as stale persistent memory. 
Even in verifier-rich environments, where task success can often be checked after execution, existing self-evolution mechanisms lack a principled criterion for assessing an update's transferability, retention value, and commitment risk. 
The central question is \textbf{how to govern the transition from temporary experience to persistent agent state} under transfer, safety, and regression constraints.

\begin{wrapfigure}{r}{0.5\linewidth}   %
    \centering
    \includegraphics[width=\linewidth]{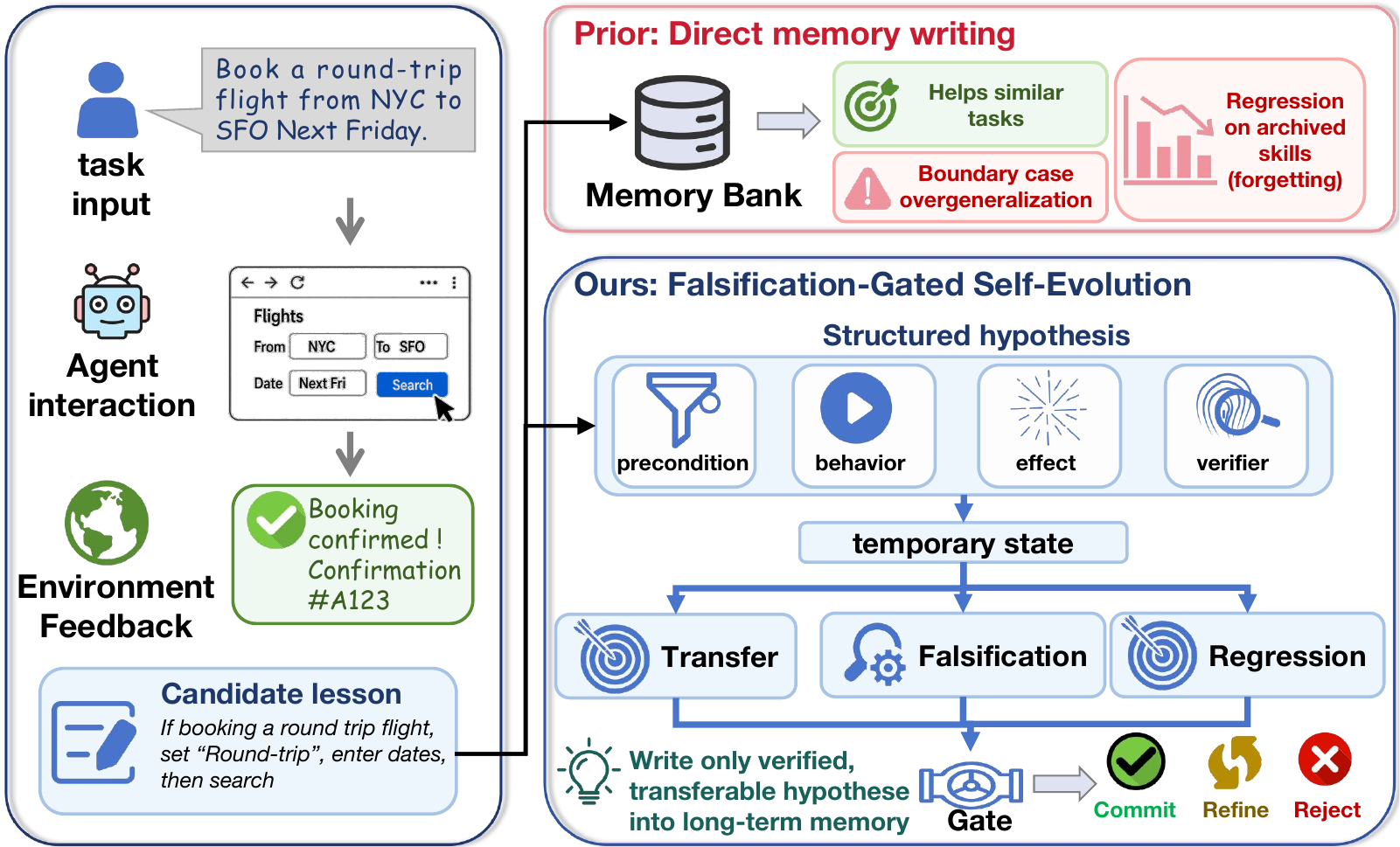}
    \caption{
        A locally useful episode-derived lesson may overgeneralize or regress archived skills if directly written into memory.
        FGSE tests each lesson as a structured hypothesis in a temporary state and commits only verified, transferable updates to long-term memory.
    }
    \label{fig:teaser}
\end{wrapfigure}

We study when an experience-derived behavioral update should enter an LLM agent's long-term memory. 
Our approach treats persistent updates as testable hypotheses. 
As shown in Fig.~\ref{fig:teaser}, our approach temporarily applies each candidate update, evaluates its transfer beyond the source episode, robustness to targeted counterexamples, and preservation of archived successful behavior, and then commits, refines, or rejects it.

We make the following contributions:
(\textbf{i}) We identify a core failure mode of self-evolving agents: target improvement alone is an unreliable signal for persistent memory writing, because many apparently useful updates encode unsafe shortcuts or overly broad applicability conditions.
(\textbf{ii}) We introduce \textbf{Falsification-Gated Self-Evolution} (FGSE), a verifier-grounded framework that treats candidate updates as testable hypotheses and performs structured commit--refine--reject triage before long-term retention.
(\textbf{iii}) Across web, app, tool-use, and long-term memory settings, FGSE improves task performance while reducing falsification failures, regression damage, harmful committed updates, and post-commit violations compared with reflection-, memory-, and context-evolution baselines.

\section{Related Work}
\label{sec:related_work}

\textbf{Interactive language agents and verifier-rich environments.}
LLM agents combine reasoning with external actions, from ReAct~\citep{yao2023react} to tool use~\citep{qin2024toolllm,chen2026semverbenchbenchmarkingllmcomprehension,xiao2026cora}, while LLM-derived traffic context also augments motion prediction~\citep{zheng2024large}. 
This progress has enabled realistic web, app, and tool-use benchmarks~\citep{zhou2024webarena,trivedi2024appworld,yao2024taubench,debenedetti2024agentdojo}, where success can often be checked by executable validators, database states, user goals, or safety constraints.
Healthcare-agent evaluation protocols additionally assess episode-level state continuity, evidence traceability, and escalation decisions~\citep{yao2026workflowawarebenchmarkinghealthcarenlp}. 
Complementary work verifies and repairs synthetic web environments before agent training~\citep{zhang2026trustworthyworlds}.
A YANG-grounded LLM agent supports multi-vendor OpenROADM management through schema retrieval and validation of generated configurations~\citep{xiao2026yang}.
Complementary network-diagnosis work uses multi-head graph attention to jointly address alarm correlation and root alarm detection in optical transport networks~\citep{xiao2025joint}.
Medical-agent security combines risk perception, evidence constraints, and knowledge-consistency verification~\citep{hu2026research}.
Broader security research studies quantitative vulnerability assessment~\citep{shao2026policy}, SHAP-based explanations for network anomaly detection~\citep{shao2026interpretable}, APT attribution over heterogeneous threat knowledge graphs~\citep{wang2026trident}, and provenance-graph anomaly detection~\citep{ma2026musk}. 
FGSE builds on verifier-rich settings and applies verification to candidate behavioral updates before persistent commitment, testing transfer, falsification, and regression safety.

\textbf{Memory and context-based self-improvement.}
Another line improves agents without updating model weights. Self-Refine~\citep{madaan2023selfrefine} and Reflexion~\citep{shinn2023reflexion} revise future behavior through self-feedback or verbal reinforcement, while Voyager~\citep{wang2023voyager} stores reusable executable skills. Persistent-memory systems such as Generative Agents~\citep{park2023generative}, MemoryBank~\citep{zhong2024memorybank}, MemGPT~\citep{packer2023memgpt}, and A-Mem~\citep{xu2025amem} organize memories for retrieval, reflection, and personalization, with LoCoMo~\citep{maharana2024locomo} and LongMemEval~\citep{wu2024longmemeval} evaluating long-horizon memory utility. Recent context-evolution methods, including Symbolic Learning~\citep{zhou2025symboliclearning}, Lighthouse~\citep{yang2025lighthouse}, and ACE~\citep{zhang2026ace}, further revise prompts, playbooks, critiques, or contexts from execution feedback. These methods show that textual updates can improve agents, but they may overgeneralize local lessons, accumulate stale or conflicting memories, or degrade useful context during iterative rewriting, echoing concerns about unreliable self-correction~\citep{huang2024cannotselfcorrect,liu2026passing} and unsafe misevolution~\citep{shao2026misevolve}. Adjacent retrieval studies distinguish semantic relevance from factual consistency in RAG~\citep{ye2026seeingconflict} and show benefits of exemplar diversity under matched context budgets for multi-turn intent understanding~\citep{lin2026beyondcontext}. FGSE treats each lesson as a structured, verifier-grounded hypothesis and commits it only after falsification and regression testing.

\textbf{Capability-level contribution evidence.}
\citet{xu2026operationalvalue} propose an observable evidence standard for added planning capabilities, separating presence, opportunity to act, executed-plan change, and incremental operational consequence under controlled replay, with explicit treatment of equivalence, uncertainty, and operating scope.
Their analysis requires capability-specific contribution evidence for acceptance, beyond aggregate policy gains or proxy diagnostics.
For memory updates, contribution likewise depends on verified behavioral effects beyond storage or retrieval.
FGSE evaluates each update against the current agent in a temporary state and requires target-transfer gain, falsification resistance, and regression preservation before commitment (Eq.~\eqref{eq:commit_rule}); subsequent activation requires a matching precondition and a violation record below threshold (Section~\ref{sec:runtime_use_main}).
These conditions link contribution evidence to verification and bounded activation of memory updates.

\textbf{Automated agent optimization and self-evolution.}
Beyond memory writing, recent work optimizes agent systems by searching over prompts, graph structures, workflows, or multi-agent state machines~\citep{yang2024opro,zhuge2024gptswarm,zhang2025aflow,hu2025adas,zhang2025metaagent}, and improves reasoning policies through test-time reinforcement learning, self-generated curricula, or tool-integrated co-evolution~\citep{zuo2025ttrl,tian2025seea,huang2026rzero,xia2025agent0}. 
In Text-to-SQL, an autonomy-based taxonomy highlights execution-guided self-correction and modular agent collaboration~\citep{su2027agenticsqltaxonomy}.
Continual adaptation also studies how to preserve knowledge under bounded resources: CoMem~\citep{zhou2026comem} uses concept-graph memory and feature-space rehearsal for vision--language adaptation; InduceKV~\citep{chen2026inducekv} externalizes multimodal adaptation into budgeted key--value memories; and Meta-UCF~\citep{xiao2026metaucf} generates task-conditioned LoRA updates with a shared hypernetwork.
Compo-ReAlign~\citep{xiao2026comporealign} preserves compositional structure through reversible alignment and sensitivity-constrained updates, while Affordance-First Decomposition~\citep{xu2026affordancefirst} separates a stable affordance substrate from query-routed, conflict-aware adaptation in video--language learning.
These lines address adaptation capacity, retention, and resource efficiency, alongside cross-domain studies of recoverable model pruning~\citep{lu2026fcos}, multi-scale and attention-based visual processing~\citep{zhang2023multi,zhang2024cf,zhang2026decoding}, activation design and LLM-guided fusion for forecasting~\citep{zhang2026rethinking,zhang2026timesaf}, and operational decision making through semantic evidence for inspection allocation~\citep{he2026semantic}, causal intervention ranking~\citep{huang2026dacri}, and accuracy--stability trade-offs in forecasting~\citep{li2026accuracypreserving}. 
FGSE addresses a complementary update-level problem: it falsifies each episode-derived update before persistent commitment, preserving self-evolution plasticity while controlling unsafe transfer and regression.

\section{Preliminary Study}
\label{sec:preliminary}

A common self-improvement strategy is to summarize completed episodes into reflective text or experiential insights and store them in memory for future decision making~\citep{shinn2023reflexion}.
While such updates can help on similar tasks, they may overgeneralize, disrupt previously solved behaviors, or encode brittle shortcuts.
We characterize these failure modes in a diagnostic study.

\textbf{Diagnostic setup.}
For each completed episode, the base agent generates a free-form update $u$, such as a rule, reminder, or behavioral correction, which we inject into the memory prompt and compare against the original agent.
For analysis, we evaluate $u$ on four case groups: source replay, near transfer, boundary contrast, and archived success, corresponding to the original episode neighborhood, intended-pattern transfer, superficially similar non-trigger cases, and previously solved tasks.
For each group $\mathcal C$, we measure the average verifier-score change induced by $u$:
\begin{equation}
\Delta_{\mathcal C}(u)
=
\frac{1}{|\mathcal C|}
\sum_{x\in \mathcal C}
\left[
J(\Theta \oplus u, x) - J(\Theta, x)
\right],
\label{eq:prelim_delta}
\end{equation}
where $\Theta$ is the original agent state and $\Theta\oplus u$ denotes the same agent with the free-form lesson $u$ appended to memory. 
We report near-transfer gain $\Delta_{\mathrm{near}}(u)$, boundary harm $[-\Delta_{\mathrm{boundary}}(u)]_+$, and archive harm $[-\Delta_{\mathrm{archive}}(u)]_+$. 
We also annotate each lesson with a coarse scope-specificity score in $[0,1]$, where lower values indicate broader and less conditional updates.

\begin{figure*}[t]
    \centering
    \includegraphics[width=\linewidth]{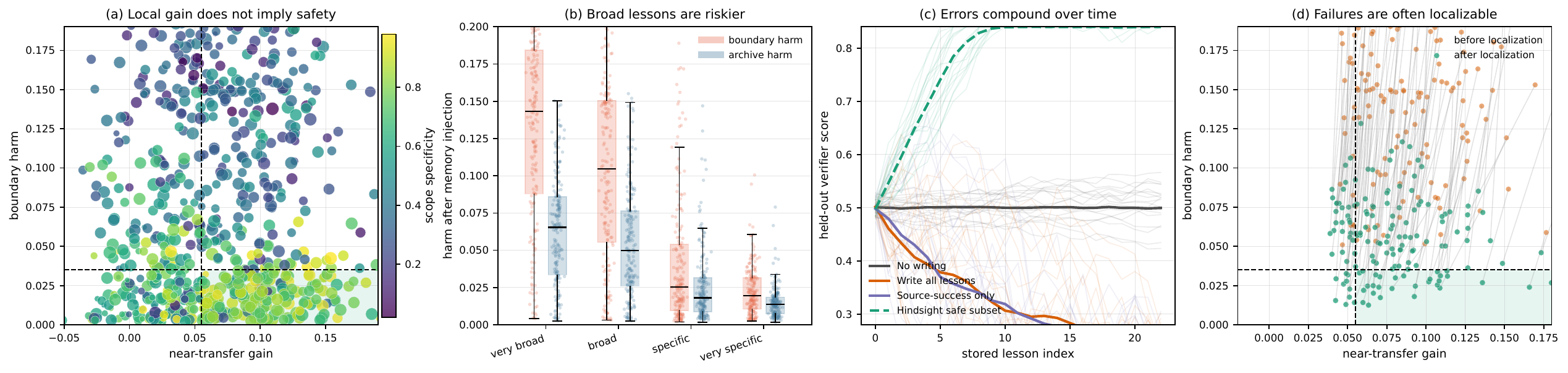}
    \caption{
    \textbf{Exploratory diagnostics for naive episode-derived self-updates.}
    \textbf{(a)} Many updates combine near-transfer gains with boundary harm. 
    \textbf{(b)} Broad and underspecified updates produce substantially higher boundary and archive harm than more specific updates. 
    \textbf{(c)} Sequentially writing free-form lessons into memory can accumulate errors, especially when updates are selected only by source-episode improvement. 
    \textbf{(d)} A post-hoc scope-localization analysis shows that many harmful but useful updates can be repaired by narrowing their applicability, motivating an explicit refinement mechanism.
    }
    \label{fig:prelim_diagnostics}
\end{figure*}

\textbf{Findings.}
Figure~\ref{fig:prelim_diagnostics} shows that naive memory writing is useful but unreliable. 
Free-form lessons often improve near-transfer cases but can cause boundary harm, indicating that locally valid behaviors may be applied in inappropriate contexts. 
This failure is closely tied to scope: broad, underspecified lessons produce larger boundary and archive degradation than more specific ones, and sequentially writing them into memory accumulates errors despite short-term source gains. 
Post-hoc localization further shows that some harmful updates contain valid rules with overly broad applicability; narrowing their scope can retain transfer benefits while reducing harm. 
These findings motivate FGSE, which treats episode-derived lessons as structured, falsifiable hypotheses, tests them temporarily on target, boundary, and archived cases, and commits or refines them only when supported by evidence.

\section{Method}
\label{sec:method}

\subsection{Overview}
\label{sec:method_overview}

We propose \textbf{Falsification-Gated Self-Evolution} (FGSE), a verifier-grounded framework for reliable long-term improvement of language agents. Given an episode, FGSE converts the observed success or failure into a structured behavioral hypothesis that specifies its precondition, behavior change, expected effect, and verifier. The hypothesis is applied only temporarily and evaluated on target-transfer probes and falsification probes. It is committed to persistent agent state only when it improves target cases, survives targeted falsification, and preserves previously solved behavior. Hypotheses with localized failures are refined once through precondition narrowing and reevaluated; hypotheses that still fail are rejected. Figure~\ref{fig:fgse_pipeline} summarizes the pipeline, with probe construction and refinement details provided in Appendix~\ref{app:probe_construction} and Appendix~\ref{app:scope_refinement}.

\begin{figure}[t]
	\centering
	\includegraphics[width=\linewidth]{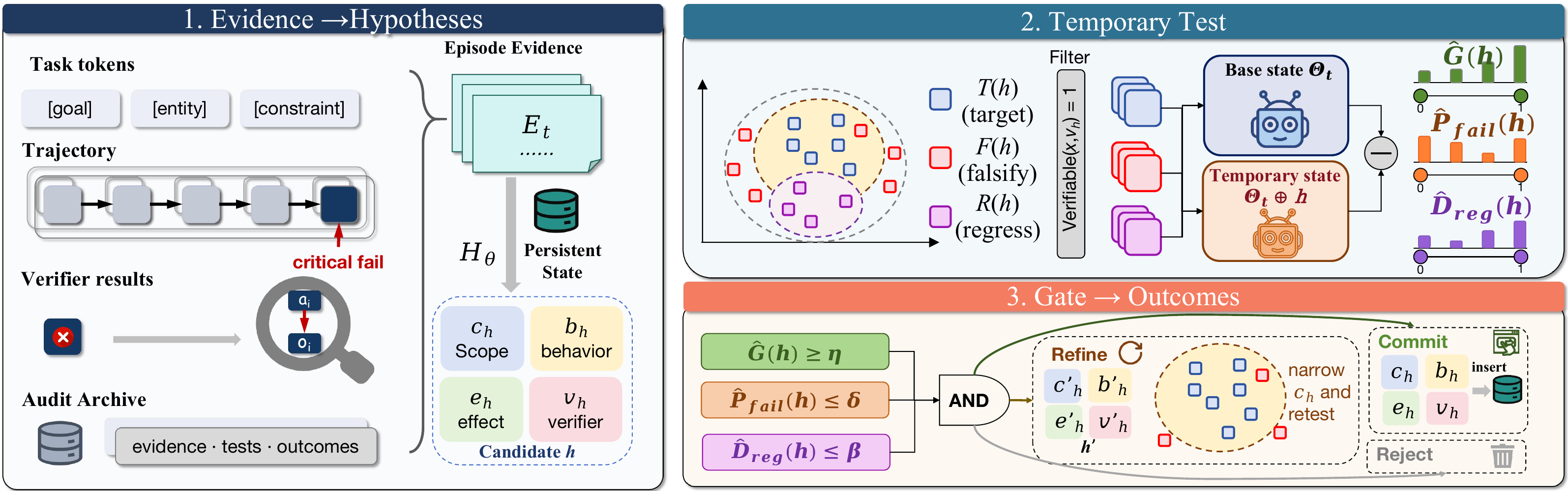}
	\caption{\textbf{Overview of Falsification-Gated Self-Evolution (FGSE).}
FGSE first converts episode evidence into a structured behavioral hypothesis,
temporarily evaluates it with target-transfer, falsification, and regression probes,
and then commits, refines, or rejects the hypothesis according to verifier-grounded
gain and risk criteria.}
	\label{fig:fgse_pipeline}
\end{figure}

\subsection{Falsification-Gated Self-Evolution}
\label{sec:fgse_main}

\paragraph{Verifier-grounded agent state.}
At episode $t$, the agent receives a task $x_t$ and acts under persistent state
\begin{equation}
\Theta_t=(\mathcal M_t,\pi_t),
\qquad
J_v(\Theta,x)
=
\mathbb E_{\tau\sim p_{\Theta}(\cdot\mid x)}
\left[v(x,\tau)\right].
\label{eq:state_and_score}
\end{equation}
Here, $\mathcal M_t$ is the retrievable memory of committed evolution units, $\pi_t$ is the current agent policy, $p_{\Theta}(\tau\mid x)$ is the trajectory distribution induced by state $\Theta$ on task $x$, and $v(x,\tau)\in[0,1]$ is a verifier-grounded score. The expectation accounts for stochastic decoding, tool uncertainty, and multi-step interaction. A completed episode yields a trajectory $\tau_t$ and verifier result $r_t=v_t(x_t,\tau_t)$, from which FGSE forms an evidence packet $E_t=(x_t,\tau_t,r_t,\mathcal A_t)$ using the audit archive $\mathcal A_t$.

\paragraph{Structured evolution hypothesis.}
Each candidate update is represented as
\begin{equation}
h=(c_h,b_h,e_h,v_h).
\label{eq:hypothesis}
\end{equation}
The precondition $c_h:\mathcal Z\rightarrow[0,1]$ assigns an applicability score to an interaction state $z=(x,\tau_{<\ell})$. The behavior transformation $b_h$ maps the current policy to a modified policy $b_h[\pi_\Theta]$. The expected-effect checker $e_h$ specifies the verifier-observable outcome that should hold after applying the update. The verifier $v_h$ provides the task-level score used to evaluate improvement, failure, and regression. The proposer $H_\phi$ maps the evidence packet to a schema-constrained hypothesis, $h_t=H_\phi(E_t)$. The proposer generates candidates; the commitment gate determines their retention.

The hypothesis is activated only when its precondition is satisfied:
\begin{equation}
g_h(z)=\mathbb I[c_h(z)\geq \gamma_h],
\qquad
\pi_{\Theta_t\oplus h}(a\mid z)
=
(1-\lambda_h g_h(z))\pi_{\Theta_t}(a\mid z)
+
\lambda_h g_h(z)b_h[\pi_{\Theta_t}](a\mid z).
\label{eq:temporary_policy}
\end{equation}
Here, $\gamma_h$ is the activation threshold, $\lambda_h\in[0,1]$ is the intervention strength, and $\Theta_t\oplus h$ denotes temporary application of $h$ without changing persistent memory. The temporary policy remains a valid action distribution because Eq.~\eqref{eq:temporary_policy} is a convex combination of two valid policies.

\paragraph{Verifier-grounded probes.}
For a candidate hypothesis $h$, FGSE constructs a target-transfer set $\mathcal T(h)$, a falsification set $\mathcal F(h)$, and a regression set $\mathcal R(h)$:
\begin{equation}
\mathcal P(h)=\mathcal T(h)\cup\mathcal F(h)\cup\mathcal R(h).
\label{eq:probe_sets}
\end{equation}
The target-transfer set contains cases where the hypothesis is expected to help. The falsification set contains cases designed to expose overgeneralization. The regression set $\mathcal R(h)$ contains archived cases that the current agent already solves and that must not be degraded. A probe $x$ is admitted only when $\operatorname{Verifiable}(x,v_h)=1$, ensuring that the gate is based on executable or externally checkable signals. FGSE builds falsification probes from boundary cases, precondition violations, regression cases, and decoys; their construction is described in Appendix~\ref{app:probe_construction}.

\paragraph{Temporary evaluation.}
FGSE evaluates $h$ by comparing the original state $\Theta_t$ with the temporary state $\Theta_t\oplus h$ on the same probes:
\begin{equation}
\Delta_h(x)
=
J_{v_h}(\Theta_t\oplus h,x)
-
J_{v_h}(\Theta_t,x).
\label{eq:update_effect}
\end{equation}
A positive $\Delta_h(x)$ means that the hypothesis improves verifier-grounded performance on $x$, while a negative value indicates harm. Empirical estimation under repeated rollouts is given in Appendix~\ref{app:empirical_estimation}.

Let $E_h(\Theta,x)\in\{0,1\}$ denote whether the expected effect $e_h$ is satisfied under state $\Theta$ on task $x$. FGSE computes
\begin{equation}
\begin{aligned}
\widehat G(h) &= \frac{1}{|\mathcal T(h)|} \sum_{x\in\mathcal T(h)} \Delta_h(x), \\
\widehat P_{\mathrm{fail}}(h) &= \frac{1}{|\mathcal F(h)|} \sum_{x\in\mathcal F(h)} \mathbb I\!\left[ \Delta_h(x)<-\epsilon \vee E_h(\Theta_t\oplus h,x)=0 \right], \\
\widehat D_{\mathrm{reg}}(h) &= \frac{1}{|\mathcal R(h)|} \sum_{x\in\mathcal R(h)} [-\Delta_h(x)]_+ .
\end{aligned}
\label{eq:evaluation_statistics}
\end{equation}
Here, $\epsilon\geq0$ is the tolerated verifier-score drop and $[\cdot]_+=\max(\cdot,0)$. The three statistics measure useful transfer, targeted falsification risk, and regression damage.

\paragraph{Commitment gate.}
The commitment decision is
\begin{equation}
\mathsf{Commit}(h)
=
\mathbb I
\left[
\widehat G(h)\geq\eta
\wedge
\widehat P_{\mathrm{fail}}(h)\leq\delta
\wedge
\widehat D_{\mathrm{reg}}(h)\leq\beta
\right],
\label{eq:commit_rule}
\end{equation}
where $\eta>0$ is the minimum target-transfer gain, $\delta\in[0,1]$ is the maximum falsification rate, and $\beta\geq0$ is the maximum regression damage. If $\widehat G(h)\geq\eta$ but failures are concentrated outside a separable sub-scope, FGSE performs one precondition-refinement pass and reevaluates the refined hypothesis. The refinement objective and separability criterion are given in Appendix~\ref{app:scope_refinement}. A hypothesis that fails after refinement is rejected.

The persistent state is updated by
\begin{equation}
\Theta_{t+1}
=
\begin{cases}
\Theta_t\oplus h, & \mathsf{Commit}(h)=1,\\
\Theta_t, & \mathsf{Commit}(h)=0.
\end{cases}
\label{eq:state_update}
\end{equation}
A committed hypothesis is stored together with its precondition, behavior change, verifier, evaluated probes, verifier outcomes, and post-commit violation record. Rejected hypotheses are excluded from retrievable memory. 

\subsection{Runtime Use and Monitoring}
\label{sec:runtime_use_main}

During future inference, FGSE retrieves committed hypotheses whose preconditions match the current interaction state and whose violation records remain below threshold. Activated hypotheses modify the policy through the same behavior-transformation interface as Eq.~\eqref{eq:temporary_policy}. When multiple hypotheses are active, FGSE combines them by normalized precondition scores while preserving a nonzero base-policy contribution. Post-commit violations update a running violation score; hypotheses exceeding the violation threshold are removed from retrieval. The retrieval, combination, and monitoring rules are provided in Appendix~\ref{app:runtime_monitoring}.

\begin{algorithm}[t]
\caption{Falsification-Gated Self-Evolution}
\label{alg:fgse}
\begin{algorithmic}[1]
\REQUIRE Agent state $\Theta_t$, task $x_t$, trajectory $\tau_t$, verifier result $r_t$, audit archive $\mathcal A_t$
\ENSURE Updated state $\Theta_{t+1}$
\STATE Form evidence packet $E_t=(x_t,\tau_t,r_t,\mathcal A_t)$.
\STATE Propose structured hypothesis $h=H_\phi(E_t)$ with fields $(c_h,b_h,e_h,v_h)$.
\STATE Construct $\mathcal T(h)$, $\mathcal F(h)$, and $\mathcal R(h)$ under verifier constraint $\operatorname{Verifiable}(x,v_h)=1$.
\STATE Temporarily apply $h$ and evaluate $\Delta_h(x)$ for all $x\in\mathcal T(h)\cup\mathcal F(h)$.
\STATE Compute $\widehat G(h)$, $\widehat P_{\mathrm{fail}}(h)$, and $\widehat D_{\mathrm{reg}}(h)$ using Eq.~\eqref{eq:evaluation_statistics}.
\IF{$\mathsf{Commit}(h)=1$}
    \STATE Set $\Theta_{t+1}=\Theta_t\oplus h$ and record the evaluated evidence.
\ELSIF{$\widehat G(h)\geq\eta$ and the failures satisfy the refinement criterion in Appendix~\ref{app:scope_refinement}}
    \STATE Refine the precondition, reevaluate the refined hypothesis, and apply Eq.~\eqref{eq:state_update}.
\ELSE
    \STATE Reject $h$ and set $\Theta_{t+1}=\Theta_t$.
\ENDIF
\end{algorithmic}
\end{algorithm}

\section{Theoretical Analysis}
\label{sec:theory}

We analyze whether a hypothesis that passes the finite-probe commitment gate is reliable on the underlying target, falsification, and regression probe distributions. 
Fix a round \(t\) and condition on the pre-evaluation history, so the current state \(\Theta_t\), evidence packet, and finite candidate set \(\mathcal H_t\) are fixed. 
If scope refinement is used, refinement is performed on a development split, while final commitment uses fresh probes.

For each \(h\in\mathcal H_t\), let \(\mathcal D_T(h)\), \(\mathcal D_F(h)\), and \(\mathcal D_R(h)\) denote the target-transfer, falsification, and regression probe distributions. 
We define
\begin{equation}
\label{eq:theory_population_quantities}
\begin{aligned}
G(h)
&=
\mathbb E_{x\sim\mathcal D_T(h)}
\left[\Delta_h(x)\right],
\\
P_{\mathrm{fail}}(h)
&=
\mathbb P_{x\sim\mathcal D_F(h)}
\left[
\Delta_h(x)<-\epsilon
\ \vee\
E_h(\Theta_t\oplus h,x)=0
\right],
\\
D_{\mathrm{reg}}(h)
&=
\mathbb E_{x\sim\mathcal D_R(h)}
\left[[-\Delta_h(x)]_+\right].
\end{aligned}
\end{equation}
These quantities measure useful transfer, falsification risk, and regression damage, respectively. 
Let \(\mathrm{LCB}_T(h)\), \(\mathrm{UCB}_F(h)\), and \(\mathrm{UCB}_R(h)\) be one-sided confidence bounds for these three quantities, computed from fresh final probes using empirical-Bernstein bounds for real-valued gains and Bernoulli-KL bounds for falsification events. 
Their finite-sample forms are given in Appendix~\ref{app:confidence_bounds}.

\begin{theorem}[Generalization of the confidence-gated commitment rule]
\label{thm:commitment_gate_generalization}
Suppose that, conditional on the pre-evaluation history, the final probes for each \(h\in\mathcal H_t\) are sampled from \(\mathcal D_T(h)\), \(\mathcal D_F(h)\), and \(\mathcal D_R(h)\). 
Then, with probability at least \(1-\alpha\), all \(h\in\mathcal H_t\) simultaneously satisfy
\begin{equation}
\label{eq:theory_confidence_event}
G(h)\ge \mathrm{LCB}_T(h),
\qquad
P_{\mathrm{fail}}(h)\le \mathrm{UCB}_F(h),
\qquad
D_{\mathrm{reg}}(h)\le \mathrm{UCB}_R(h).
\end{equation}
Consequently, any hypothesis satisfying
\begin{equation}
\label{eq:theory_confidence_gate}
\mathrm{LCB}_T(h)\ge \eta,
\qquad
\mathrm{UCB}_F(h)\le \delta,
\qquad
\mathrm{UCB}_R(h)\le \beta
\end{equation}
also satisfies the population constraints
\begin{equation}
\label{eq:theory_population_safety}
G(h)\ge \eta,
\qquad
P_{\mathrm{fail}}(h)\le \delta,
\qquad
D_{\mathrm{reg}}(h)\le \beta.
\end{equation}
\end{theorem}

\paragraph{Proof sketch.}
Conditioning on the pre-evaluation history fixes \(\mathcal H_t\) and prevents final probes from being reused for hypothesis construction. 
Empirical-Bernstein inequalities control the bounded target-transfer and regression quantities, while a Bernoulli-KL bound controls falsification risk. 
A union bound over hypotheses and statistics gives the simultaneous high-probability event, under which the confidence-gated rule implies the corresponding population constraints.

\begin{corollary}[Probe complexity of safe commitment]
\label{cor:probe_complexity_safe_commitment}
Up to logarithmic factors in \(|\mathcal H_t|\), \(1/\alpha\), and the number of falsification probes, it suffices to use
\begin{equation}
\label{eq:theory_probe_complexity_rate}
n_T=\widetilde O\!\left(\frac{1}{\zeta_T^2}\right),
\qquad
n_R=\widetilde O\!\left(\frac{1}{\zeta_R^2}\right),
\qquad
n_F=\widetilde O\!\left(
\frac{1}{\mathrm{kl}(\delta_-\|\delta)}
\right)
\end{equation}
final target, regression, and falsification probes to ensure that any hypothesis with empirical target margin \(\zeta_T\), empirical regression margin \(\zeta_R\), and empirical falsification rate at most \(\delta_-<\delta\) satisfies \(G(h)\ge\eta\), \(P_{\mathrm{fail}}(h)\le\delta\), and \(D_{\mathrm{reg}}(h)\le\beta\) with probability at least \(1-\alpha\). 
The exact finite-sample statement is proved in Appendix~\ref{app:proof_probe_complexity_corollary}.
\end{corollary}

\begin{table*}[t]
\centering
\scriptsize
\caption{
\textbf{Main comparison on web, app, and tool-interaction benchmarks.}
For AppWorld, Avg. is the mean of Test-Normal and Test-Challenge task-goal completion.
Best and second-best scores in each column are shown in \textbf{bold} and \underline{underlined}, respectively.
}
\label{tab:main_agent_results}
\resizebox{\textwidth}{!}{
\begin{tabular}{llcccccc}
\toprule
\textbf{Method}
& \textbf{Backbone / Setting}
& \makecell{\textbf{WebArena}\\\textbf{SR} $\uparrow$}
& \makecell{\textbf{AppWorld}\\\textbf{Normal} $\uparrow$}
& \makecell{\textbf{AppWorld}\\\textbf{Challenge} $\uparrow$}
& \makecell{\textbf{AppWorld}\\\textbf{Avg.} $\uparrow$}
& \makecell{\textbf{$\tau$-Retail}\\\textbf{Pass$^1$} $\uparrow$}
& \makecell{\textbf{$\tau$-Airline}\\\textbf{Pass$^1$} $\uparrow$} \\
\midrule

\grouprow{Web / browser-control agents: methods primarily designed or evaluated for browser navigation and web interaction.}

GPT-4 Agent~\citep{zhou2024webarena}
& GPT-4
& 14.4
& 43.2
& 25.4
& 34.3
& 0.553
& 0.378 \\

AutoEval~\citep{pan2024autoeval}
& GPT-4
& 20.2
& 45.8
& 28.7
& 37.3
& 0.579
& 0.391 \\

BrowserGym~\citep{chezelles2025browsergym}
& GPT-4o
& 24.3
& 46.5
& 29.1
& 37.8
& 0.585
& 0.402 \\

CER$_{\mathrm{online}}$~\citep{liu2025cer}
& GPT-4o
& 33.2
& 51.9
& 33.4
& 42.7
& 0.622
& 0.424 \\

CER$_{\mathrm{offline}}$~\citep{liu2025cer}
& GPT-4o
& 33.4
& 52.6
& 34.1
& 43.4
& 0.631
& 0.429 \\

CER$_{\mathrm{hybrid}}$~\citep{liu2025cer}
& GPT-4o
& 36.7
& 53.0
& 35.5
& 44.3
& 0.638
& 0.437 \\

\midrule
\grouprow{App / API-interaction agents: methods primarily evaluated on app-level tool use, database-state goals, and API execution.}

IPFunCall~\citep{trivedi2024appworld}
& GPT-4o
& 11.8
& 32.1
& 18.0
& 25.1
& 0.482
& 0.309 \\

FullCodeRefl~\citep{trivedi2024appworld}
& GPT-4o
& 12.7
& 33.9
& 19.2
& 26.6
& 0.497
& 0.318 \\

PlanExec~\citep{trivedi2024appworld}
& GPT-4o
& 15.9
& 44.6
& 19.7
& 32.2
& 0.534
& 0.347 \\

ReAct~\citep{yao2023react,trivedi2024appworld}
& GPT-4o
& 18.1
& 48.8
& 30.2
& 39.5
& 0.558
& 0.325 \\

ReAct + ICL~\citep{zhang2026ace}
& DeepSeek-V3.1
& 19.6
& 64.3
& 46.0
& 55.2
& 0.615
& 0.408 \\

ReAct + GEPA~\citep{agrawal2026gepa,zhang2026ace}
& DeepSeek-V3.1
& 20.1
& 64.9
& 46.0
& 55.5
& 0.620
& 0.415 \\

ReAct + DC~\citep{suzgun2025dc,zhang2026ace}
& DeepSeek-V3.1
& 20.6
& 65.5
& 52.3
& 58.9
& 0.630
& 0.421 \\

ReAct + ACE$_{\mathrm{offline}}$~\citep{zhang2026ace}
& DeepSeek-V3.1
& 25.7
& 76.2
& 57.3
& 66.8
& 0.654
& 0.448 \\

ReAct + ACE$_{\mathrm{online}}$~\citep{zhang2026ace}
& DeepSeek-V3.1
& 26.4
& 69.6
& 66.7
& 68.2
& 0.665
& 0.456 \\

\midrule
\grouprow{Tool-user interaction agents: methods centered on multi-turn user simulation, tool calling, and domain-policy execution.}

Tool-Calling~\citep{yao2024taubench}
& GPT-4o
& 17.3
& 50.4
& 31.6
& 41.0
& 0.604
& 0.420 \\

Tool-Calling~\citep{yao2024taubench}
& Claude-3.5-Sonnet
& 18.6
& 54.2
& 34.7
& 44.5
& 0.692
& 0.460 \\

Act~\citep{yao2024taubench}
& GPT-4o
& 15.8
& 47.7
& 28.2
& 38.0
& 0.545
& 0.365 \\

ReAct~\citep{yao2023react,yao2024taubench}
& GPT-4o
& 16.4
& 48.1
& 29.0
& 38.6
& 0.521
& 0.325 \\

\midrule
\grouprow{Long-term memory agents: methods primarily designed for persistent memory, conversational memory, or memory-augmented retrieval.}

Offline Reading~\citep{wu2024longmemeval}
& GPT-4o oracle
& 3.5
& 12.4
& 6.3
& 9.4
& 0.180
& 0.121 \\

Coze~\citep{wu2024longmemeval}
& GPT-4o
& 9.4
& 24.0
& 12.9
& 18.5
& 0.410
& 0.270 \\

ChatGPT~\citep{wu2024longmemeval}
& GPT-4o
& 13.8
& 43.0
& 26.0
& 34.5
& 0.560
& 0.390 \\

GPT-4o + Chain-of-Note~\citep{wu2024longmemeval}
& long-context
& 14.6
& 45.0
& 28.5
& 36.8
& 0.575
& 0.399 \\

ChatGPT~\citep{wu2024longmemeval}
& GPT-4o-mini
& 12.2
& 36.5
& 20.9
& 28.7
& 0.520
& 0.348 \\

ReadAgent~\citep{xu2025amem}
& GPT-4o-mini
& 7.4
& 25.3
& 12.6
& 19.0
& 0.420
& 0.260 \\

MemoryBank~\citep{xu2025amem}
& GPT-4o-mini
& 6.8
& 22.9
& 10.4
& 16.7
& 0.390
& 0.247 \\

MemGPT~\citep{xu2025amem}
& GPT-4o-mini
& 10.7
& 29.7
& 16.2
& 23.0
& 0.450
& 0.286 \\

LoCoMo~\citep{maharana2024locomo,xu2025amem}
& GPT-4o-mini
& 11.5
& 31.6
& 17.3
& 24.5
& 0.463
& 0.300 \\

A-Mem~\citep{xu2025amem}
& GPT-4o-mini
& 14.9
& 38.2
& 22.5
& 30.4
& 0.544
& 0.368 \\

\midrule

\rowcolor{oursblue}
FGSE (Ours)
& GPT-4o-mini
& 38.6 $\pm$ 1.8
& 82.7 $\pm$ 2.2
& 66.1 $\pm$ 2.0
& 74.4 $\pm$ 1.6
& 0.707 $\pm$ 0.015
& \second{0.478} $\pm$ 0.012 \\

\rowcolor{oursblue}
FGSE (Ours)
& GPT-4o
& \second{39.2} $\pm$ 1.5
& \second{84.8} $\pm$ 1.3
& \second{68.3} $\pm$ 1.8
& \second{76.6} $\pm$ 1.8
& \second{0.716} $\pm$ 0.012
& 0.471 $\pm$ 0.010 \\

\rowcolor{oursblue}
FGSE (Ours)
& DeepSeek-V3.1
& \best{39.8} $\pm$ 1.2
& \best{88.2} $\pm$ 1.5
& \best{69.4} $\pm$ 1.0
& \best{78.8} $\pm$ 1.7
& \best{0.724} $\pm$ 0.010
& \best{0.487} $\pm$ 0.008 \\

\bottomrule
\end{tabular}
}
\vspace{-1mm}
\end{table*}

\section{Experiments}
\subsection{Experimental Setup}
\label{sec:exp_setup}

\textbf{Datasets.}
We evaluate FGSE on verifier-grounded agentic tasks where updates can be tested before persistent commitment. 
The benchmark suite covers web and browser control with WebArena~\citep{zhou2024webarena} and BrowserGym~\citep{chezelles2025browsergym}, app and API interaction with AppWorld~\citep{trivedi2024appworld} and $\tau$-bench~\citep{yao2024taubench}, adversarial tool-use robustness with AgentDojo~\citep{debenedetti2024agentdojo}, and long-term memory reliability with LongMemEval~\citep{wu2024longmemeval} and LoCoMo~\citep{maharana2024locomo}. 
We also include verifier-rich coding and reasoning diagnostics on HumanEval~\citep{chen2021codex}, MBPP~\citep{austin2021program}, SWE-bench Verified~\citep{jimenez2024swebench,chowdhury2024swebenchverified}, GSM8K~\citep{cobbe2021gsm8k}, and MATH~\citep{hendrycks2021math}. 
Details are provided in App.~\ref{app:exp_details}.

\textbf{Evaluation metrics.}
We report each benchmark's official task metric, including success rate, task-goal completion, pass rate or $\mathrm{pass}^k$, utility and security success, QA accuracy/F1, pass@1, resolved rate, and exact-match accuracy. 
To evaluate safe self-evolution, we additionally report target-transfer gain $\widehat G$, falsification failure rate $\widehat P_{\mathrm{fail}}$, regression damage $\widehat D_{\mathrm{reg}}$, post-commit violation rate, and harmful update rate.

\begin{wrapfigure}{R}{0.6\textwidth}  %
    \centering
    \includegraphics[width=\linewidth]{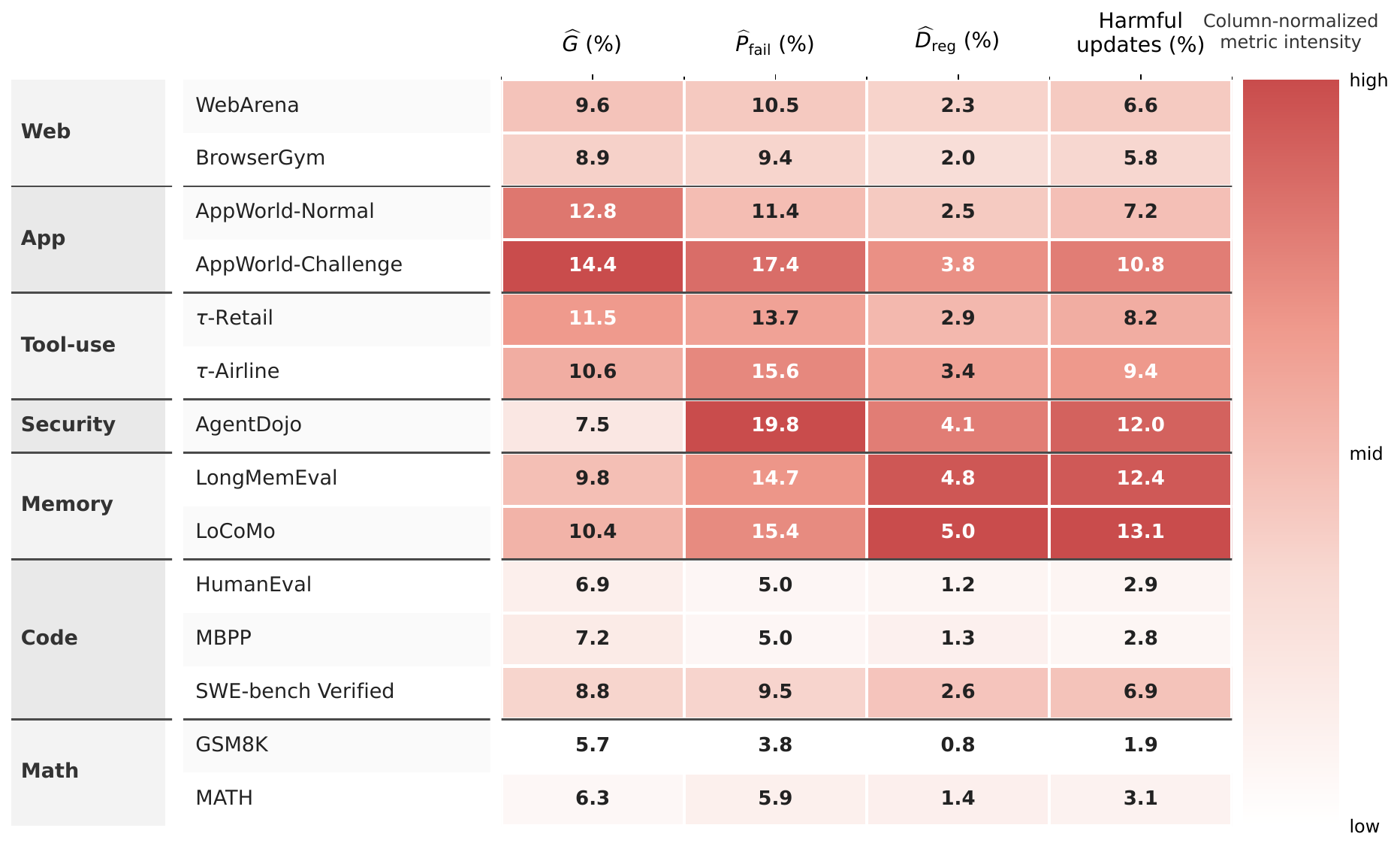}
    \caption{
    \textbf{Domain-level diagnostic of self-evolution risk.}
    }
    \label{fig:domain_metric_heatmap}
\end{wrapfigure}

\textbf{Baselines.}
We compare FGSE with non-evolving agents, including direct prompting and ReAct~\citep{yao2023react}, as well as self-update baselines such as Self-Refine~\citep{madaan2023selfrefine}, Reflexion~\citep{shinn2023reflexion}, Voyager-style skill memory~\citep{wang2023voyager}, and naive episodic memory. 
We further compare with recent self-evolving or agent-optimization methods, including Symbolic Learning~\citep{zhou2025symboliclearning}, AlphaLLM~\citep{tian2024alphallm}, GPTSwarm~\citep{zhuge2024gptswarm}, AFlow~\citep{zhang2025aflow}, ADAS~\citep{hu2025adas}, MetaAgent~\citep{zhang2025metaagent}, A-Mem~\citep{xu2025amem}, Lighthouse~\citep{yang2025lighthouse}, and ACE~\citep{zhang2026ace}. 
For methods requiring workflow search, external data, or model updates, we separate controlled comparisons from native-setting results, since FGSE modifies only the persistent agent state and commitment rule.

\textbf{Implementation details.}
The default backbone is a ReAct-style executor with persistent memory retrieval. 
FGSE proposes at most $B=3$ candidate hypotheses per episode and evaluates each using target-transfer, falsification, and regression probes with default budgets $|\mathcal T(h)|=8$, $|\mathcal F(h)|=16$, and $|\mathcal R(h)|=16$. 
We tune only $\eta$, $\delta$, and $\beta$ on the development split, using defaults $\eta=0.03$, $\delta=0.05$, $\beta=0.01$, $\epsilon=0$, $\gamma_h=0.5$, and $\lambda_h=0.8$. 
Stochastic settings use three random seeds, and we report mean performance with bootstrap confidence intervals. Refer to App.~\ref{app:exp_details} for more details.

\subsection{Main Results}
\label{sec:main_results}

\paragraph{Overall comparison.}
Table~\ref{tab:main_agent_results} shows that FGSE improves over memory-writing
and reflection-style baselines across verifier-rich web, app, tool-use, and memory
benchmarks.
The gains are especially clear on verifier-rich settings such as WebArena, AppWorld, and $\tau$-bench, where tool-state and database-state validators provide reliable evidence for pre-commitment testing. 
Appendix~\ref{app.meo} further shows that FGSE improves long-term memory reliability while reducing falsification failures, regression damage, and harmful committed updates.

\textbf{Domain-level update diagnostics.}
Beyond task success, Figure~\ref{fig:domain_metric_heatmap} reports update-centric metrics across domains, including $\widehat G$, $\widehat P_{\mathrm{fail}}$, $\widehat D_{\mathrm{reg}}$, and harmful update rate. 
AppWorld, $\tau$-bench, AgentDojo, and long-term memory benchmarks show both meaningful gains and higher safety risks, indicating greater vulnerability to overgeneralized self-updates. 
Coding and math tasks exhibit lower falsification and regression risks due to cleaner executable or exact-match verifiers, while SWE-bench remains regression-prone because repository-level changes are highly contextual. 
These results support verifier-grounded falsification over commitment based only on target improvement.

\begin{figure*}[t]
    \centering
    \begin{minipage}[t]{0.565\textwidth}
        \centering
        \includegraphics[width=\linewidth]{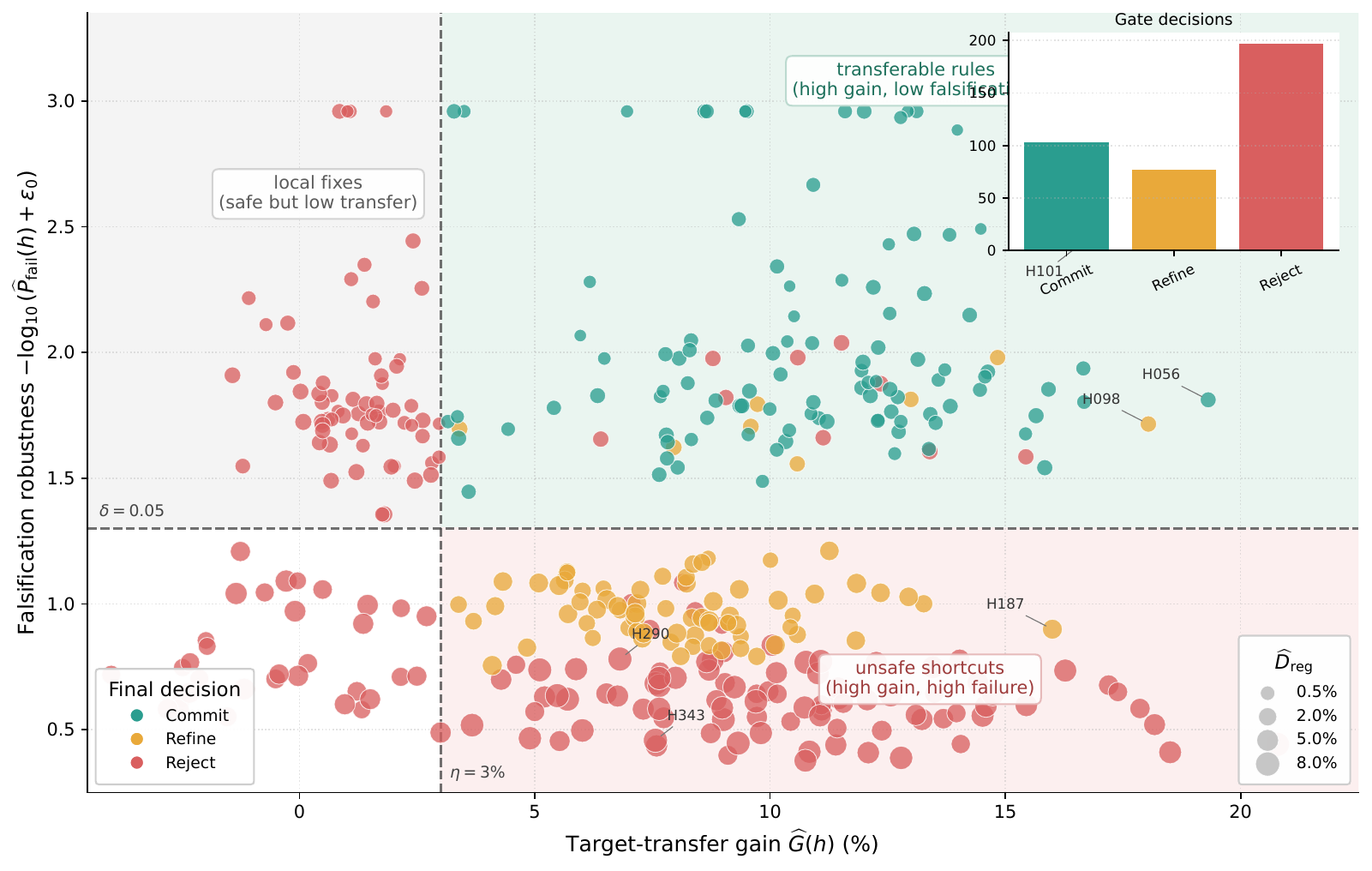}
        \vspace{-1mm}
        {\small \textbf{(a)} Transfer--Falsification profile}
    \end{minipage}
    \hfill
    \begin{minipage}[t]{0.405\textwidth}
        \centering
        \includegraphics[width=\linewidth]{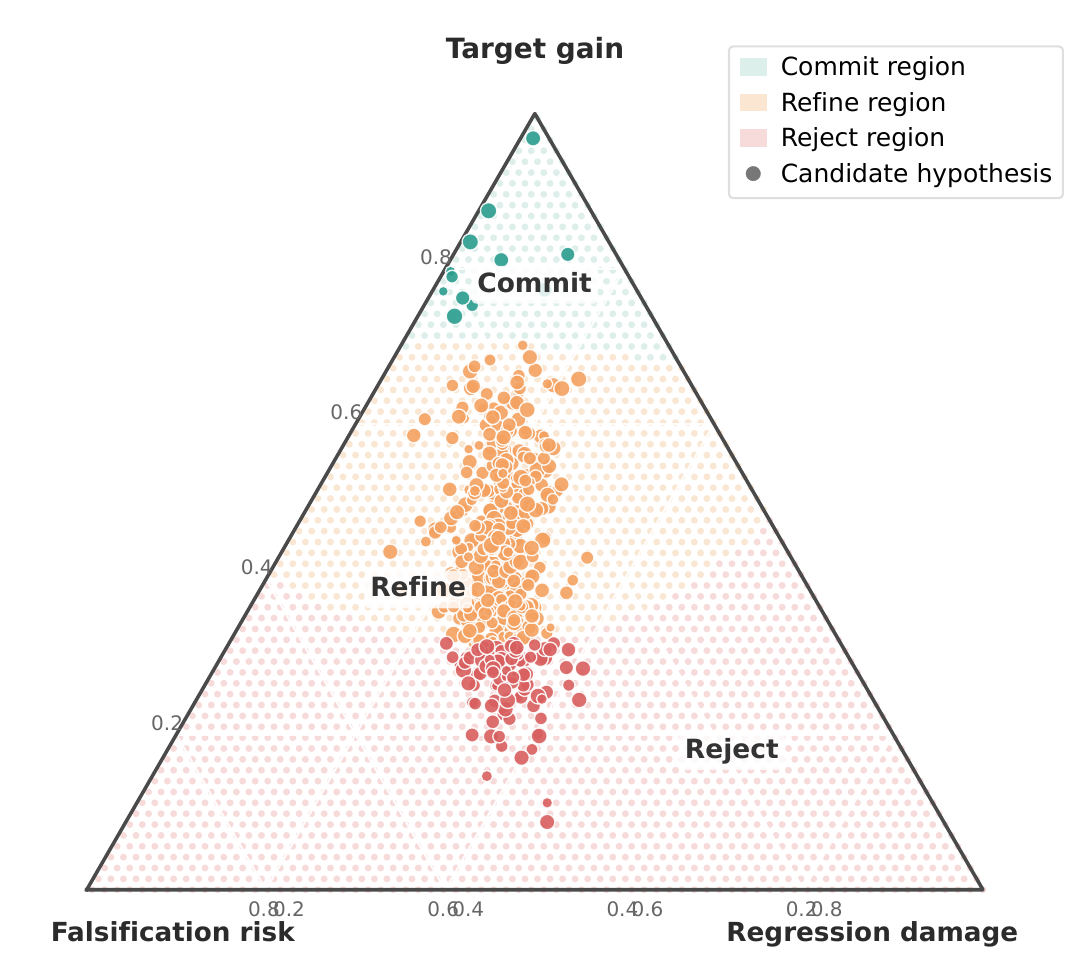}
        \vspace{-1mm}
        {\small \textbf{(b)} Decision simplex}
    \end{minipage}
    \caption{
    \textbf{Interpreting FGSE update selection.}
    \textbf{(a)} Transfer--Falsification volcano plot for candidate hypotheses. The $x$-axis shows target-transfer gain $\widehat G$, the $y$-axis shows $-\log_{10}(\widehat P_{\mathrm{fail}}+\epsilon_0)$, point size encodes regression damage $\widehat D_{\mathrm{reg}}$, and color indicates the final FGSE decision. The dashed lines mark the useful-transfer and falsification thresholds.
    \textbf{(b)} Decision simplex over normalized target gain, falsification risk, and regression damage. The shaded area shows the gate-induced decision regions. %
    }
    \label{fig:fgse_gate_diagnostics}
    \label{fig:transfer_falsification_volcano}
    \label{fig:decision_simplex}
\end{figure*}

\subsection{Ablation and Mechanism Analysis}
\label{sec:ablation}

\paragraph{Single-factor ablation.}
We isolate FGSE components by removing or replacing one component at a time and report both task performance and update reliability. 
Blue values in parentheses denote absolute changes from full FGSE, with negative values indicating drops for higher-is-better metrics and positive values indicating worse lower-is-better metrics. 
Tables~\ref{tab:ablation_task_performance} and~\ref{tab:ablation_update_reliability} show that each component contributes. 
Removing structured hypotheses causes the largest drop, reducing WebArena SR by $4.2$ points, AppWorld Avg. by $5.9$ points, and LoCoMo Avg. R-L by $4.78$ points, confirming the need for scoped, verifier-aware updates. 
Target-transfer probes improve useful generalization, as $\widehat G$ drops from $12.9\%$ to $9.8\%$ without them, while falsification probes prevent unsafe overgeneralization, with $\widehat P_{\mathrm{fail}}$ increasing from $5.0\%$ to $18.6\%$ and harmful updates from $3.6\%$ to $10.6\%$ when removed. 
The regression archive reduces $\widehat D_{\mathrm{reg}}$ from $5.2\%$ to $1.3\%$, and scope refinement, post-commit monitoring, and executable verifiers further improve the transfer-safety trade-off.

\begin{table*}[t]
\centering
\scriptsize
\caption{
\textbf{Single-factor ablation on task-level performance.}
}
\label{tab:ablation_task_performance}
\begin{tabularx}{\textwidth}{l *{5}{>{\centering\arraybackslash}X}}
\toprule
\textbf{Variant}
& \makecell{\textbf{WebArena}\\\textbf{SR} $\uparrow$}
& \makecell{\textbf{AppWorld}\\\textbf{Avg.} $\uparrow$}
& \makecell{\textbf{$\tau$-Retail}\\\textbf{Pass$^1$} $\uparrow$}
& \makecell{\textbf{LongMemEval}\\\textbf{Acc.} $\uparrow$}
& \makecell{\textbf{LoCoMo}\\\textbf{Avg. R-L} $\uparrow$} \\
\midrule

\rowcolor{oursblue}
Full FGSE
& \best{39.8}
& \best{78.8}
& \best{0.724}
& \best{0.748}
& \best{37.62} \\

\quad w/o structured hypothesis
& 35.6\abldrop{-4.2}
& 72.9\abldrop{-5.9}
& 0.664\abldrop{-0.060}
& 0.692\abldrop{-0.056}
& 32.84\abldrop{-4.78} \\

\quad w/o target-transfer probes
& 36.8\abldrop{-3.0}
& 74.9\abldrop{-3.9}
& 0.684\abldrop{-0.040}
& 0.710\abldrop{-0.038}
& 34.72\abldrop{-2.90} \\

\quad w/o falsification probes
& 37.4\abldrop{-2.4}
& 75.8\abldrop{-3.0}
& 0.697\abldrop{-0.027}
& 0.713\abldrop{-0.035}
& 35.10\abldrop{-2.52} \\

\quad w/o regression archive
& 38.2\abldrop{-1.6}
& 76.9\abldrop{-1.9}
& 0.706\abldrop{-0.018}
& 0.719\abldrop{-0.029}
& 35.87\abldrop{-1.75} \\

\quad w/o scope refinement
& 37.1\abldrop{-2.7}
& 75.4\abldrop{-3.4}
& 0.690\abldrop{-0.034}
& 0.716\abldrop{-0.032}
& 34.96\abldrop{-2.66} \\

\quad LLM-judge gate
& 36.7\abldrop{-3.1}
& 74.1\abldrop{-4.7}
& 0.681\abldrop{-0.043}
& 0.707\abldrop{-0.041}
& 34.18\abldrop{-3.44} \\

\quad w/o post-commit monitoring
& \second{39.0}\abldrop{-0.8}
& \second{77.2}\abldrop{-1.6}
& \second{0.713}\abldrop{-0.011}
& \second{0.726}\abldrop{-0.022}
& \second{35.92}\abldrop{-1.70} \\

\bottomrule
\end{tabularx}
\vspace{-1mm}
\end{table*}

\paragraph{Transferability of candidate updates.}
We assess whether episode-derived updates capture reusable rules that transfer beyond the triggering case. 
We therefore evaluate each candidate before commitment using three temporary statistics: target-transfer gain $\widehat G(h)$, falsification failure rate $\widehat P_{\mathrm{fail}}(h)$, and regression damage $\widehat D_{\mathrm{reg}}(h)$. 
Figure~\ref{fig:fgse_gate_diagnostics}(a) plots candidates in a Transfer--Falsification volcano plot, with $\widehat G(h)$ on the $x$-axis, $-\log_{10}(\widehat P_{\mathrm{fail}}(h)+\epsilon_0)$ on the $y$-axis, point size encoding $\widehat D_{\mathrm{reg}}(h)$, and color indicating the gate decision. 
The plot reveals three groups: committed updates in the upper right with high transfer and low risk, unsafe shortcuts in the lower right, and safe but low-transfer local fixes in the upper left. 
In this run, FGSE commits $104/380$ candidates, achieving average $\widehat G=10.77\%$, $\widehat P_{\mathrm{fail}}=1.28\%$, and $\widehat D_{\mathrm{reg}}=0.60\%$. 
The $101$ rejected candidates that exceed the gain threshold $\eta$ show much higher falsification and regression risk, with $\widehat P_{\mathrm{fail}}=22.82\%$ and $\widehat D_{\mathrm{reg}}=4.85\%$. 
Refined candidates lie between these groups, suggesting that many useful but unsafe updates require scope narrowing before commitment.

\textbf{Commit--refine--reject decision boundary.}
To examine the gate, we map each candidate into a simplex over normalized target gain, falsification risk, and regression damage. 
For each candidate \(h\), we normalize the three statistics across candidates and project them onto simplex coordinates \((g_h,f_h,r_h)\) with \(g_h+f_h+r_h=1\). 
Figure~\ref{fig:fgse_gate_diagnostics}(b) shows the joint role of the three statistics: \textsc{commit} lies in the high-gain, low-risk region, \textsc{reject} concentrates near the high-risk, high-regression edge, and \textsc{refine} forms a transition band between them. 
This pattern indicates that FGSE performs structured triage, committing gain-dominant updates, refining repairable but ambiguous ones, and rejecting risk-dominant candidates.

\section{Conclusion}

We studied reliable self-evolution for long-horizon LLM agents, focusing on when episode-derived lessons should enter persistent memory. 
FGSE treats each candidate update as a structured, verifier-grounded hypothesis and commits it only after target-transfer, falsification, and regression testing. 
Our results show that safe adaptation requires evaluating usefulness, boundary failures, and regression on previously solved behavior alongside source-task improvement. 
Future work can extend falsification-gated update control to multimodal agents~\citep{xu2026lever,dong2026imageslookrightretrieve,liu2026dualpathwaycircuitsobjecthallucination,zhang-etal-2026-modality,kang2026drilltrainingworldmodels}, collaborative tool-use systems~\citep{fan2026communicationaware}, and settings where verifiers must be learned or approximated.

\begin{ack}
We thank Airon Tech for providing computing resources and data-processing support for this work.
The authors declare no competing interests.
\end{ack}

\bibliographystyle{plainnat}
\bibliography{ref}

\newpage

\appendix

\begin{center}
\small
\renewcommand{\arraystretch}{1.18}
\setlength{\tabcolsep}{8pt}
\arrayrulecolor{appLine}

\begin{tabularx}{0.98\linewidth}{A B}
\toprule
\rowcolor{appLight}
\textbf{\textcolor{appBlue}{Appendix}}
&
\textbf{\textcolor{appBlue}{Second-Level Headings}} \\
\midrule

\appSection{app:technical_details}{Additional Technical Details}
&
\appSub{app:prelim_details}{Details of the Exploratory Preliminary Study}
\appSub{app:probe_construction}{Verifier-Grounded Probe Construction}
\appSub{app:scope_refinement}{Precondition Refinement}
\appSub{app:empirical_estimation}{Empirical Estimation Under Stochastic Rollouts}
\appSub{app:runtime_monitoring}{Runtime Retrieval, Combination, and Monitoring}
\appSub{app:audit_record}{Audit Record}
\appSub{app:detailed_algorithm}{Detailed Algorithm with Refinement}
\\[0.25em]

\midrule

\appSection{app:commitment_gate_proofs}{Proofs for the Commitment Gate Generalization}
&
\appSub{app:conditional_setup}{Conditional Setup}
\appSub{app:confidence_bounds}{Confidence Bounds Used by the Commitment Gate}
\appSub{app:concentration_tools}{Concentration Tools}
\appSub{app:uniform_confidence_event}{Uniform Confidence Event}
\appSub{app:proof_commitment_gate_theorem}{Proof of Theorem~\ref*{thm:commitment_gate_generalization}}
\appSub{app:exact_probe_complexity}{Exact Probe Complexity Statement}
\appSub{app:proof_probe_complexity_corollary}{Proof of Corollary~\ref*{cor:probe_complexity_safe_commitment}}
\appSub{app:adaptivity_and_refinement}{Remarks on Adaptivity and Scope Refinement}
\\[0.25em]

\midrule

\appSection{app:exp_details}{Additional Experimental Details}
&
\appSub{app:benchmark_details}{Benchmark Details}
\appSub{app:metrics_details}{Evaluation Metrics}
\appSub{app:fair_comparison}{Fair Comparison Protocol}
\appSub{app:baseline_details}{Baseline Implementation Details}
\appSub{app:fgse_hyperparameters}{FGSE Hyperparameters}
\appSub{app:ablations}{Ablation Studies}
\appSub{app:compute_resources}{Compute Resources}
\\[0.25em]

\midrule

\appSection{app:additional_results}{Additional Experimental Results}
&
\appSub{app:reliability_ablation}{Additional Reliability and Ablation Results}
\appSub{app:update_selection_analysis}{Update-Selection and Falsification Analysis}
\appSub{app:mechanistic_analysis}{Mechanistic Analysis of Long-Term Self-Evolution}
\appSub{app:sensitivity_scalability}{Cost and Scalability}
\\[0.25em]

\midrule

\appSection{app:broader_impacts}{Broader Impacts}
&
\vspace{1pt}
\textcolor{appGray}{-}
\\[0.25em]

\midrule

\appSection{app:llm_usage}{LLM Usage}
&
\vspace{1pt}
\textcolor{appGray}{-}
\\

\bottomrule
\end{tabularx}

\arrayrulecolor{black}
\end{center}

\vspace{0.4em}
\section{Additional Technical Details}
\label{app:technical_details}

\subsection{Details of the Exploratory Preliminary Study}
\label{app:prelim_details}

The preliminary study in Section~\ref{sec:preliminary} examines \emph{naive episode-derived memory writing}. 
The agent directly summarizes a completed episode into a free-form lesson and writes it into persistent memory.

\paragraph{Episode-derived lessons.}
For each completed episode, the base agent observes the task instruction $x$, its interaction trajectory $\tau$, and the verifier outcome $r$. 
It then produces a natural-language lesson
\begin{equation}
u = \mathrm{SummarizeLesson}(x,\tau,r),
\label{eq:app_lesson_generation}
\end{equation}
which may take the form of a rule, reminder, correction, or behavioral heuristic. 
The updated agent state is obtained by simply appending $u$ to the memory context:
\begin{equation}
\Theta \oplus u
=
(\mathcal M \cup \{u\}, \pi),
\label{eq:app_naive_memory_write}
\end{equation}
where $\mathcal M$ denotes the current memory and $\pi$ denotes the unchanged base policy. 
This direct memory-writing procedure is a common self-improvement baseline that exposes unsafe updates.

\paragraph{Probe groups.}
To analyze the effect of a lesson $u$, we evaluate the original agent $\Theta$ and the memory-augmented agent $\Theta\oplus u$ on four groups of diagnostic cases. 
The \emph{source-replay} group $\mathcal C_{\mathrm{src}}(u)$ contains the original episode and minor perturbations that preserve the same intended solution. 
The \emph{near-transfer} group $\mathcal C_{\mathrm{near}}(u)$ contains tasks that share the same underlying behavioral pattern but differ in surface form, entities, or context. 
The \emph{boundary-contrast} group $\mathcal C_{\mathrm{bdry}}(u)$ contains cases that are superficially similar to the source episode but should not trigger the same lesson. 
The \emph{archived-success} group $\mathcal C_{\mathrm{arch}}(u)$ contains previously solved cases sampled from the agent's historical archive. 
The first two groups measure whether the lesson captures a useful pattern, while the latter two groups measure whether it overgeneralizes or interferes with existing behavior.

\paragraph{Verifier-score changes.}
For a case $x$, we estimate the verifier-grounded performance of an agent state $\Theta$ as
\begin{equation}
J(\Theta,x)
=
\frac{1}{L}\sum_{\ell=1}^{L} v(x,\tau^{(\ell)}),
\qquad
\tau^{(\ell)}\sim p_{\Theta}(\cdot\mid x),
\label{eq:app_prelim_score}
\end{equation}
where $v(x,\tau)\in[0,1]$ is the task verifier and $L$ is the number of repeated rollouts. 
For deterministic tasks, $L=1$ is sufficient. 
The effect of lesson $u$ on a case group $\mathcal C$ is
\begin{equation}
\Delta_{\mathcal C}(u)
=
\frac{1}{|\mathcal C|}
\sum_{x\in\mathcal C}
\left[
J(\Theta\oplus u,x)-J(\Theta,x)
\right].
\label{eq:app_group_delta}
\end{equation}
We report source gain $\Delta_{\mathrm{src}}(u)$, near-transfer gain $\Delta_{\mathrm{near}}(u)$, boundary harm
\begin{equation}
H_{\mathrm{bdry}}(u)
=
\left[-\Delta_{\mathcal C_{\mathrm{bdry}}}(u)\right]_+,
\label{eq:app_boundary_harm}
\end{equation}
and archive harm
\begin{equation}
H_{\mathrm{arch}}(u)
=
\left[-\Delta_{\mathcal C_{\mathrm{arch}}}(u)\right]_+.
\label{eq:app_archive_harm}
\end{equation}
A lesson is considered locally useful if it improves near-transfer cases, but it is considered unsafe if it also induces large boundary or archive harm.

\paragraph{Scope-specificity annotation.}
To understand why some useful lessons are harmful, we annotate each free-form lesson with a coarse scope-specificity score $s(u)\in[0,1]$. 
A lesson receives a higher score when it contains explicit applicability conditions, contextual triggers, exception clauses, or task-specific constraints. 
It receives a lower score when it is phrased as a broad heuristic without clear conditions. 
This analysis-only score tests whether overgeneralization is associated with underspecified lesson scope; update selection in the preliminary study proceeds without it.

\paragraph{Sequential memory-writing analysis.}
To study error accumulation, we simulate sequential memory writing over a stream of candidate lessons. 
After writing $k$ lessons, the agent state is
\begin{equation}
\Theta_k
=
\Theta_0 \oplus \{u_1,\ldots,u_k\},
\label{eq:app_seq_state}
\end{equation}
and its held-out verifier score is measured on a fixed evaluation set $\mathcal E$:
\begin{equation}
S_k
=
\frac{1}{|\mathcal E|}
\sum_{x\in\mathcal E}
J(\Theta_k,x).
\label{eq:app_seq_score}
\end{equation}
We compare three diagnostic memory-writing strategies: writing all lessons, writing only lessons that improve the source episode, and a hindsight safe subset. 
The hindsight subset keeps lessons with high near-transfer gain and low boundary/archive harm, and is used only as an oracle diagnostic upper bound. 
It is not available to the agent during deployment. 
This comparison isolates the main observation that source improvement alone is insufficient for stable long-term self-evolution.

\paragraph{Post-hoc scope localization.}
For lessons that are useful but harmful, we conduct a post-hoc localization analysis. 
Given a lesson $u$ with positive near-transfer gain but high boundary harm, we inspect the boundary failures and rewrite the lesson into a narrower version $u'$ by adding an explicit applicability condition. 
We then reevaluate $u'$ on the same near-transfer and boundary-contrast groups. 
We summarize the effect of localization by the transfer-retention ratio
\begin{equation}
R_{\mathrm{retain}}(u')
=
\frac{\Delta_{\mathrm{near}}(u')}
{\Delta_{\mathrm{near}}(u)+\varepsilon},
\label{eq:app_transfer_retention}
\end{equation}
and the boundary-harm reduction ratio
\begin{equation}
R_{\mathrm{reduce}}(u')
=
\frac{
H_{\mathrm{bdry}}(u)-H_{\mathrm{bdry}}(u')
}{
H_{\mathrm{bdry}}(u)+\varepsilon
},
\label{eq:app_harm_reduction}
\end{equation}
where $\varepsilon$ is a small constant for numerical stability. 
This separate diagnostic motivates explicit applicability conditions and refinement of useful but overly broad candidates in the main method.

\paragraph{Connection to the method.}
The preliminary study supports three design choices. 
First, testing episode-derived updates before persistence can expose unsafe free-form lessons. 
Second, contrastive and archived cases expose boundary failures that source or near-transfer cases can miss. 
Third, scope refinement can repair some harmful lessons while retaining their useful content. 
These observations motivate the structured hypothesis representation, temporary evaluation, verifier-grounded commitment, and scope refinement introduced in Section~\ref{sec:method}.

\subsection{Verifier-Grounded Probe Construction}
\label{app:probe_construction}

FGSE constructs probes from three candidate sources: archived solved tasks, verifier-preserving perturbations of the triggering episode, and language-model-generated candidates filtered by executable verifiers. The verifier filter is applied before evaluation:
\begin{equation}
\operatorname{Verifiable}(x,v_h)=1
\quad\Longleftrightarrow\quad
v_h(x,\tau)\ \text{returns a stable score for every evaluated rollout } \tau .
\label{eq:verifiable_filter}
\end{equation}
Tasks that cannot be checked by $v_h$ are excluded from the gate.

The falsification set is decomposed as
\begin{equation}
\mathcal F(h)
=
\mathcal F_{\mathrm{bd}}(h)
\cup
\mathcal F_{\mathrm{pv}}(h)
\cup
\mathcal F_{\mathrm{rg}}(h)
\cup
\mathcal F_{\mathrm{dc}}(h).
\label{eq:falsification_decomposition_appendix}
\end{equation}
Here, $\mathcal F_{\mathrm{bd}}$ contains boundary probes near the activation threshold of $c_h$, $\mathcal F_{\mathrm{pv}}$ contains precondition-violation probes that are similar to the claimed scope but should not activate the rule, $\mathcal F_{\mathrm{rg}}$ contains regression probes from previously solved cases, and $\mathcal F_{\mathrm{dc}}$ contains decoy probes that are superficially similar to target cases but require different behavior.

For each probe type $q\in\{\mathrm{bd},\mathrm{pv},\mathrm{rg},\mathrm{dc}\}$, FGSE assigns a risk score
\begin{equation}
\rho_q(x;h)
=
\begin{cases}
\exp\!\left(-|c_h(x)-\gamma_h|/\sigma\right),
& q=\mathrm{bd},\\[2pt]
(1-c_h(x))\,\operatorname{sim}(x,h),
& q=\mathrm{pv},\\[2pt]
\mathbb I[J_{v_h}(\Theta_t,x)\geq \zeta]\,\operatorname{sim}(x,h),
& q=\mathrm{rg},\\[2pt]
\operatorname{sim}(x,h)\,\mathbb I[c_h(x)<\gamma_h],
& q=\mathrm{dc}.
\end{cases}
\label{eq:risk_score_appendix}
\end{equation}
In Eq.~\eqref{eq:risk_score_appendix}, $\sigma>0$ controls the width of the boundary region, $\operatorname{sim}(x,h)$ measures semantic or structural similarity between $x$ and the triggering context of $h$, and $\zeta$ is the score threshold for considering an archived case solved.

The selected probes are
\begin{equation}
\mathcal F_q(h)
=
\operatorname{TopK}_{x\in\mathcal X_q}
\left\{
\rho_q(x;h):
\operatorname{Verifiable}(x,v_h)=1
\right\},
\label{eq:topk_probe_selection_appendix}
\end{equation}
where $\mathcal X_q$ is the candidate pool for probe type $q$. Boundary probes test whether the hypothesis changes behavior too aggressively near its activation threshold. Precondition-violation probes test whether the rule fires outside its intended condition. Regression probes measure damage to cases already solved by $\Theta_t$. Decoy probes test whether superficial similarity causes the wrong behavior change.

The target-transfer set $\mathcal T(h)$ is selected from verifiable candidates satisfying $c_h(x)\geq\gamma_h$. FGSE prioritizes candidates that match the precondition while differing from the triggering task in surface form, tool trace, or problem instance. This tests transfer beyond memorization of the original episode.

\subsection{Precondition Refinement}
\label{app:scope_refinement}

Refinement is used when a hypothesis improves target probes but fails on falsification probes that can be separated by a more precise precondition. FGSE forms the safe and falsifying sets
\begin{align}
\mathcal P_h^+
&=
\{x\in\mathcal T(h)\cup\mathcal F(h):\Delta_h(x)\geq0,\ E_h(\Theta_t\oplus h,x)=1\},\notag\\
\mathcal P_h^-
&=
\{x\in\mathcal F(h):\Delta_h(x)<-\epsilon\ \vee\ E_h(\Theta_t\oplus h,x)=0\}.
\label{eq:safe_and_negative_sets}
\end{align}
The refined precondition is obtained by
\begin{equation}
c_h'
=
\arg\min_{c\in\mathfrak C}
\left[
\Omega(c)
+
\lambda_+
\sum_{x\in\mathcal P_h^+}
\ell(1,c(x))
+
\lambda_-
\sum_{x\in\mathcal P_h^-}
\ell(0,c(x))
\right].
\label{eq:refinement_objective}
\end{equation}
Here, $\mathfrak C$ is the precondition class, $\Omega(c)$ penalizes overly complex conditions, $\ell$ is binary cross-entropy, and $\lambda_+,\lambda_-\geq0$ weight safe and falsifying probes.

FGSE triggers refinement only when the following separability score is below a threshold:
\begin{equation}
\operatorname{Sep}(h)
=
\min_{c\in\mathfrak C}
\left[
\frac{1}{|\mathcal P_h^+|}
\sum_{x\in\mathcal P_h^+}
\mathbb I[c(x)<\gamma_h]
+
\frac{1}{|\mathcal P_h^-|}
\sum_{x\in\mathcal P_h^-}
\mathbb I[c(x)\geq\gamma_h]
\right]
\leq \alpha_{\mathrm{sep}}.
\label{eq:separability_criterion}
\end{equation}
A refined hypothesis is $h'=(c_h',b_h,e_h,v_h)$. FGSE performs one refinement pass. The refined hypothesis is then evaluated by the same equations used in the main method. If it fails the gate after refinement, it is rejected.

\subsection{Empirical Estimation Under Stochastic Rollouts}
\label{app:empirical_estimation}

The verifier-grounded score in Eq.~\eqref{eq:state_and_score} is estimated with repeated rollouts:
\begin{equation}
\widehat J_v(\Theta,x)
=
\frac{1}{n}
\sum_{j=1}^{n}
v(x,\tau^{(j)}),
\qquad
\tau^{(j)}\sim p_{\Theta}(\cdot\mid x).
\label{eq:empirical_score_appendix}
\end{equation}
The empirical update effect is
\begin{equation}
\widehat\Delta_h(x)
=
\widehat J_{v_h}(\Theta_t\oplus h,x)
-
\widehat J_{v_h}(\Theta_t,x).
\label{eq:empirical_delta_appendix}
\end{equation}
In experiments with deterministic verifiers and deterministic decoding, $n=1$ is sufficient. With stochastic decoding or nondeterministic tool responses, FGSE uses paired random seeds for $\Theta_t$ and $\Theta_t\oplus h$ when the environment supports seed control. This reduces variance in $\widehat\Delta_h(x)$ because both states are evaluated under matched rollout conditions.

For finite probe budgets, FGSE reports the conservative falsification score
\begin{equation}
S_{\mathrm{fail}}(h)
=
\widehat P_{\mathrm{fail}}(h)
+
\kappa |\mathcal F(h)|^{-1/2},
\label{eq:conservative_failure_score}
\end{equation}
where $\kappa\geq0$ controls the penalty for small falsification sets. This score is used for analysis and ablations. The main gate in Eq.~\eqref{eq:commit_rule} uses the empirical falsification rate $\widehat P_{\mathrm{fail}}(h)$.

\subsection{Runtime Retrieval, Combination, and Monitoring}
\label{app:runtime_monitoring}

At inference time, FGSE retrieves committed hypotheses whose preconditions match the current interaction state $z=(x,\tau_{<\ell})$:
\begin{equation}
\mathcal H_t(z)
=
\{h\in\mathcal M_t:c_h(z)\geq\gamma_h,\ u_t(h)\leq u_{\max}\},
\label{eq:active_hypotheses_appendix}
\end{equation}
where $u_t(h)$ is the post-commit violation score and $u_{\max}$ is the maximum allowed violation level.

When $\mathcal H_t(z)$ is nonempty, FGSE assigns normalized weights
\begin{equation}
w_h(z)
=
\frac{\exp(c_h(z)/\rho)}
{\sum_{h'\in\mathcal H_t(z)}\exp(c_{h'}(z)/\rho)},
\qquad h\in\mathcal H_t(z),
\label{eq:runtime_weights_appendix}
\end{equation}
where $\rho>0$ is the temperature. The runtime policy is
\begin{equation}
\pi_t^{\mathrm{FGSE}}(a\mid z)
=
(1-\lambda(z))\pi_{\Theta_t}(a\mid z)
+
\lambda(z)
\sum_{h\in\mathcal H_t(z)}
w_h(z)b_h[\pi_{\Theta_t}](a\mid z),
\label{eq:runtime_policy_appendix}
\end{equation}
where $\lambda(z)\in[0,1]$ is the total intervention strength. This form preserves a nonzero base-policy contribution even when several hypotheses are active.

After a committed hypothesis is activated, FGSE records whether its expected effect is violated:
\begin{equation}
V_t(h,x_t)
=
\mathbb I
\left[
E_h(\Theta_t,x_t)=0
\vee
J_{v_h}(\Theta_t,x_t)<\zeta_h
\right],
\label{eq:runtime_violation_appendix}
\end{equation}
where $\zeta_h$ is the minimum acceptable verifier score associated with hypothesis $h$. The violation score is updated by
\begin{equation}
u_{t+1}(h)
=
\rho_u u_t(h)
+
(1-\rho_u)V_t(h,x_t),
\label{eq:violation_ema_appendix}
\end{equation}
where $\rho_u\in[0,1)$ controls the memory of previous violations. If $u_{t+1}(h)>u_{\max}$, the hypothesis is removed from the active retrieval set. Removed hypotheses remain in the audit archive and are not used for future action selection.

\subsection{Audit Record}
\label{app:audit_record}

Each committed hypothesis is stored with an audit record
\begin{equation}
\operatorname{Record}(h)
=
\left(
h,
\mathcal T(h),
\mathcal F(h),
\mathcal R(h),
\{\Delta_h(x)\}_{x\in\mathcal P(h)},
\widehat G(h),
\widehat P_{\mathrm{fail}}(h),
\widehat D_{\mathrm{reg}}(h),
u_t(h)
\right).
\label{eq:audit_record_appendix}
\end{equation}
The audit record supports post-hoc inspection of why a rule was committed, which probes were used to falsify it, and whether the rule later accumulated violations.

\subsection{Detailed Algorithm with Refinement}
\label{app:detailed_algorithm}

\begin{algorithm}[ht]
\caption{FGSE with One-Pass Precondition Refinement}
\label{alg:fgse_refinement}
\begin{algorithmic}[1]
\REQUIRE State $\Theta_t$, evidence packet $E_t$, audit archive $\mathcal A_t$, thresholds $(\eta,\delta,\beta,\alpha_{\mathrm{sep}})$
\ENSURE Updated state $\Theta_{t+1}$
\STATE Propose $h=(c_h,b_h,e_h,v_h)=H_\phi(E_t)$.
\STATE Build $\mathcal T(h)$, $\mathcal F(h)$, and $\mathcal R(h)$.
\STATE Evaluate $\widehat G(h)$, $\widehat P_{\mathrm{fail}}(h)$, and $\widehat D_{\mathrm{reg}}(h)$.
\IF{$\widehat G(h)\geq\eta$ and $\widehat P_{\mathrm{fail}}(h)\leq\delta$ and $\widehat D_{\mathrm{reg}}(h)\leq\beta$}
    \STATE Commit $h$ and set $\Theta_{t+1}=\Theta_t\oplus h$.
\ELSIF{$\widehat G(h)\geq\eta$ and $\operatorname{Sep}(h)\leq\alpha_{\mathrm{sep}}$}
    \STATE Compute $c_h'$ by Eq.~\eqref{eq:refinement_objective} and set $h'=(c_h',b_h,e_h,v_h)$.
    \STATE Rebuild probes for $h'$ and reevaluate the gate.
    \IF{$h'$ satisfies Eq.~\eqref{eq:commit_rule}}
        \STATE Commit $h'$ and set $\Theta_{t+1}=\Theta_t\oplus h'$.
    \ELSE
        \STATE Reject $h'$ and set $\Theta_{t+1}=\Theta_t$.
    \ENDIF
\ELSE
    \STATE Reject $h$ and set $\Theta_{t+1}=\Theta_t$.
\ENDIF
\end{algorithmic}
\end{algorithm}

\paragraph{Practical representation of hypotheses.}
In implementation, each hypothesis is stored as a schema-constrained JSON object.
The precondition field is a natural-language predicate together with a scoring
function implemented by a lightweight LLM classifier or rule-based matcher.
The behavior field specifies a bounded prompt/tool-policy patch. The expected-effect field specifies a verifier-observable condition,
and the verifier field points to the benchmark validator or executable checker.
Audits of synthetic healthcare workflows also expose escalation errors under conflicting evidence and brittle routing rules under unseen paraphrases~\citep{zheng2026auditingescalationreliabilityhealthcare}.

\begin{tcolorbox}[
    colback = gray!5,          %
    colframe = blue!50!black,  %
    title = JSON example,     %
    boxrule = 0.5pt,           %
    arc = 3pt,                 %
]
\begin{lstlisting}
{
  "precondition": "The task asks for a round-trip flight booking with explicit origin, destination, and dates.",
  "behavior": "Before searching, set trip_type='round_trip' and fill both departure and return dates.",
  "expected_effect": "Search query contains both outbound and return dates and trip_type='round_trip'.",
  "verifier": "booking_database_state_validator"
}
\end{lstlisting}
\end{tcolorbox}

\section{Proofs for the Commitment Gate Generalization}
\label{app:commitment_gate_proofs}

\subsection{Conditional Setup}
\label{app:conditional_setup}

We prove Theorem~\ref{thm:commitment_gate_generalization} and Corollary~\ref{cor:probe_complexity_safe_commitment}. Fix a round \(t\). Let \(\mathcal F_t\) denote the sigma-field containing the current agent state \(\Theta_t\), the evidence packet, the audit archive, the proposal and refinement randomness before final evaluation, and the finite candidate set \(\mathcal H_t\). Throughout the proof, we condition on \(\mathcal F_t\). Thus, the candidate set is fixed before the final evaluation probes are sampled.

For each \(h\in\mathcal H_t\), define the target-transfer random variable
\begin{equation}
\label{eq:app_target_random_variable}
X^T_h
=
\Delta_h(x),
\qquad
x\sim \mathcal D_T(h).
\end{equation}
Because verifier scores lie in \([0,1]\), we have
\begin{equation}
\label{eq:app_target_range}
X^T_h\in[-1,1].
\end{equation}
The population and empirical target-transfer gains are
\begin{equation}
\label{eq:app_target_mean_empirical}
G(h)
=
\mathbb E[X^T_h],
\qquad
\widehat G(h)
=
\frac{1}{n_T}
\sum_{i=1}^{n_T}
X^T_{h,i}.
\end{equation}
The empirical target-transfer variance is
\begin{equation}
\label{eq:app_target_empirical_variance}
\widehat V_T(h)
=
\frac{1}{n_T}
\sum_{i=1}^{n_T}
\left(
X^T_{h,i}
-
\widehat G(h)
\right)^2.
\end{equation}

For falsification probes, define the Bernoulli random variable
\begin{equation}
\label{eq:app_failure_random_variable}
X^F_h
=
\mathbb I
\left[
\Delta_h(x)<-\epsilon
\ \vee\
E_h(\Theta_t\oplus h,x)=0
\right],
\qquad
x\sim \mathcal D_F(h).
\end{equation}
Then
\begin{equation}
\label{eq:app_failure_range}
X^F_h\in\{0,1\}.
\end{equation}
The population falsification risk and empirical failure rate are
\begin{equation}
\label{eq:app_failure_mean_empirical}
P_{\mathrm{fail}}(h)
=
\mathbb E[X^F_h],
\qquad
\widehat P_{\mathrm{fail}}(h)
=
\frac{1}{n_F}
\sum_{i=1}^{n_F}
X^F_{h,i}.
\end{equation}

For regression probes, define
\begin{equation}
\label{eq:app_regression_random_variable}
X^R_h
=
[-\Delta_h(x)]_+,
\qquad
x\sim \mathcal D_R(h).
\end{equation}
Since \(\Delta_h(x)\in[-1,1]\), we have
\begin{equation}
\label{eq:app_regression_range}
X^R_h\in[0,1].
\end{equation}
The population and empirical regression damages are
\begin{equation}
\label{eq:app_regression_mean_empirical}
D_{\mathrm{reg}}(h)
=
\mathbb E[X^R_h],
\qquad
\widehat D_{\mathrm{reg}}(h)
=
\frac{1}{n_R}
\sum_{i=1}^{n_R}
X^R_{h,i}.
\end{equation}
The empirical regression variance is
\begin{equation}
\label{eq:app_regression_empirical_variance}
\widehat V_R(h)
=
\frac{1}{n_R}
\sum_{i=1}^{n_R}
\left(
X^R_{h,i}
-
\widehat D_{\mathrm{reg}}(h)
\right)^2.
\end{equation}

The analysis treats the randomness in \(X^T_h\), \(X^F_h\), and \(X^R_h\) as including both probe sampling and any rollout randomness used to estimate verifier outcomes. The only properties needed are boundedness and conditional independence of the final evaluation samples for each fixed \(h\).

\subsection{Confidence Bounds Used by the Commitment Gate}
\label{app:confidence_bounds}

Let \(n_T,n_F,n_R\) be the numbers of final target, falsification, and regression probes. Let \(\alpha_T,\alpha_F,\alpha_R>0\) satisfy
\begin{equation}
\label{eq:app_confidence_budget_sum}
\alpha_T+\alpha_F+\alpha_R=\alpha.
\end{equation}
Define the logarithmic factors
\begin{equation}
\label{eq:app_confidence_log_factors}
L_T
=
\log\frac{2|\mathcal H_t|}{\alpha_T},
\qquad
L_R
=
\log\frac{2|\mathcal H_t|}{\alpha_R},
\qquad
L_F
=
\log\frac{(n_F+1)|\mathcal H_t|}{\alpha_F}.
\end{equation}
The empirical-Bernstein slacks are
\begin{equation}
\label{eq:app_confidence_eb_slacks}
\varepsilon_T(h)
=
\sqrt{
\frac{2\widehat V_T(h)L_T}{n_T}
}
+
\frac{14L_T}{3(n_T-1)},
\qquad
\varepsilon_R(h)
=
\sqrt{
\frac{2\widehat V_R(h)L_R}{n_R}
}
+
\frac{7L_R}{3(n_R-1)}.
\end{equation}
The target lower confidence bound and regression upper confidence bound are
\begin{equation}
\label{eq:app_lcb_ucb_definitions}
\mathrm{LCB}_T(h)
=
\widehat G(h)-\varepsilon_T(h),
\qquad
\mathrm{UCB}_R(h)
=
\widehat D_{\mathrm{reg}}(h)+\varepsilon_R(h).
\end{equation}
For \(p,q\in[0,1]\), define the binary KL divergence
\begin{equation}
\label{eq:app_binary_kl_definition}
\mathrm{kl}(p\|q)
=
p\log\frac{p}{q}
+
(1-p)\log\frac{1-p}{1-q},
\end{equation}
with the standard continuous extension at the boundary. The KL upper confidence bound for falsification risk is
\begin{equation}
\label{eq:app_kl_ucb_definition}
\mathrm{UCB}_F(h)
=
\sup
\left\{
q\in[\widehat P_{\mathrm{fail}}(h),1]:
\mathrm{kl}
\left(
\widehat P_{\mathrm{fail}}(h)\|q
\right)
\le
\frac{L_F}{n_F}
\right\}.
\end{equation}

\subsection{Concentration Tools}
\label{app:concentration_tools}

We use two concentration results. The first controls bounded real-valued random variables using empirical variance.

\begin{lemma}[One-sided empirical Bernstein inequality]
\label{lem:app_empirical_bernstein}
Let \(Y_1,\ldots,Y_n\) be independent random variables taking values in an interval of length \(\mathfrak b\). Let
\begin{equation}
\label{eq:app_eb_mean_def}
\mu
=
\mathbb E[Y_1],
\qquad
\widehat\mu
=
\frac{1}{n}
\sum_{i=1}^{n}
Y_i,
\end{equation}
and
\begin{equation}
\label{eq:app_eb_variance_def}
\widehat V
=
\frac{1}{n}
\sum_{i=1}^{n}
(Y_i-\widehat\mu)^2.
\end{equation}
For any \(\rho\in(0,1)\) and \(n\ge2\), with probability at least \(1-\rho\),
\begin{equation}
\label{eq:app_eb_upper_tail}
\mu-\widehat\mu
\le
\sqrt{
\frac{2\widehat V\log(2/\rho)}{n}
}
+
\frac{
7\mathfrak b\log(2/\rho)
}{
3(n-1)
}.
\end{equation}
The same inequality also holds with \(\widehat\mu-\mu\) on the left-hand side.
\end{lemma}

\begin{proof}
\label{proof:app_empirical_bernstein}
This is the standard one-sided form of the empirical Bernstein inequality for bounded random variables. The lower-tail statement follows by applying the same result to \(-Y_i\), whose range length remains \(\mathfrak b\).
\end{proof}

The second result gives a KL confidence bound for Bernoulli variables.

\begin{lemma}[Bernoulli KL confidence bound]
\label{lem:app_bernoulli_kl}
Let \(B_1,\ldots,B_n\) be i.i.d. Bernoulli random variables with mean \(p\), and let
\begin{equation}
\label{eq:app_bernoulli_empirical_mean}
\widehat p
=
\frac{1}{n}
\sum_{i=1}^{n}
B_i.
\end{equation}
Then, for any \(u>0\),
\begin{equation}
\label{eq:app_bernoulli_kl_tail}
\mathbb P
\left(
\mathrm{kl}(\widehat p\|p)\ge u
\right)
\le
(n+1)\exp(-nu).
\end{equation}
\end{lemma}

\begin{proof}
\label{proof:app_bernoulli_kl}
For each \(k\in\{0,1,\ldots,n\}\), let
\begin{equation}
\label{eq:app_type_qk_def}
q_k
=
\frac{k}{n}.
\end{equation}
The event \(\widehat p=q_k\) has probability
\begin{equation}
\label{eq:app_type_probability_start}
\mathbb P(\widehat p=q_k)
=
\binom{n}{k}p^k(1-p)^{n-k}.
\end{equation}
Using the type-counting bound
\begin{equation}
\label{eq:app_type_counting_bound}
\binom{n}{k}
\le
\exp\left(nH(q_k)\right),
\end{equation}
where
\begin{equation}
\label{eq:app_entropy_definition}
H(q)
=
-q\log q-(1-q)\log(1-q),
\end{equation}
we obtain
\begin{equation}
\label{eq:app_type_probability_chain}
\begin{aligned}
\mathbb P(\widehat p=q_k)
&=
\binom{n}{k}
p^{nq_k}
(1-p)^{n(1-q_k)}
\\
&\le
\exp
\left(
nH(q_k)
+
nq_k\log p
+
n(1-q_k)\log(1-p)
\right)
\\
&=
\exp
\left(
-n
\left[
q_k\log\frac{q_k}{p}
+
(1-q_k)\log\frac{1-q_k}{1-p}
\right]
\right)
\\
&=
\exp
\left(
-n\mathrm{kl}(q_k\|p)
\right).
\end{aligned}
\end{equation}
Therefore,
\begin{equation}
\label{eq:app_bernoulli_kl_union_chain}
\begin{aligned}
\mathbb P
\left(
\mathrm{kl}(\widehat p\|p)\ge u
\right)
&=
\sum_{k:\mathrm{kl}(q_k\|p)\ge u}
\mathbb P(\widehat p=q_k)
\\
&\le
\sum_{k:\mathrm{kl}(q_k\|p)\ge u}
\exp
\left(
-n\mathrm{kl}(q_k\|p)
\right)
\\
&\le
\sum_{k=0}^{n}
\exp(-nu)
\\
&=
(n+1)\exp(-nu).
\end{aligned}
\end{equation}
This proves the claim.
\end{proof}

\subsection{Uniform Confidence Event}
\label{app:uniform_confidence_event}

We now prove that the three confidence bounds hold simultaneously for all hypotheses in \(\mathcal H_t\).

\begin{lemma}[Uniform confidence event]
\label{lem:app_uniform_confidence}
With probability at least \(1-\alpha\), every \(h\in\mathcal H_t\) simultaneously satisfies
\begin{equation}
\label{eq:app_uniform_confidence_bounds}
G(h)\ge \mathrm{LCB}_T(h),
\qquad
P_{\mathrm{fail}}(h)\le \mathrm{UCB}_F(h),
\qquad
D_{\mathrm{reg}}(h)\le \mathrm{UCB}_R(h).
\end{equation}
\end{lemma}

\begin{proof}
\label{proof:app_uniform_confidence}
We prove the target, regression, and falsification bounds separately and then apply a union bound.

First consider target-transfer gain. For a fixed \(h\), apply Lemma~\ref{lem:app_empirical_bernstein} to \(X^T_h\). Its range length is
\begin{equation}
\label{eq:app_target_range_length}
\mathfrak b_T=2.
\end{equation}
Set
\begin{equation}
\label{eq:app_target_rho_choice}
\rho_T
=
\frac{\alpha_T}{|\mathcal H_t|}.
\end{equation}
Then
\begin{equation}
\label{eq:app_target_log_conversion}
\log\frac{2}{\rho_T}
=
\log\frac{2|\mathcal H_t|}{\alpha_T}
=
L_T.
\end{equation}
For this fixed \(h\), Lemma~\ref{lem:app_empirical_bernstein} gives
\begin{equation}
\label{eq:app_target_concentration_chain}
\begin{aligned}
G(h)-\widehat G(h)
&\ge
-
\left[
\sqrt{
\frac{2\widehat V_T(h)L_T}{n_T}
}
+
\frac{7\mathfrak b_TL_T}{3(n_T-1)}
\right]
\\
&=
-
\left[
\sqrt{
\frac{2\widehat V_T(h)L_T}{n_T}
}
+
\frac{14L_T}{3(n_T-1)}
\right]
\\
&=
-\varepsilon_T(h).
\end{aligned}
\end{equation}
Thus,
\begin{equation}
\label{eq:app_target_lcb_result}
G(h)
\ge
\widehat G(h)-\varepsilon_T(h)
=
\mathrm{LCB}_T(h).
\end{equation}
A union bound over \(h\in\mathcal H_t\) gives
\begin{equation}
\label{eq:app_target_union_bound}
\mathbb P
\left(
\forall h\in\mathcal H_t:
G(h)
\ge
\mathrm{LCB}_T(h)
\right)
\ge
1-\alpha_T.
\end{equation}

Next consider regression damage. For a fixed \(h\), apply Lemma~\ref{lem:app_empirical_bernstein} to \(X^R_h\). Its range length is
\begin{equation}
\label{eq:app_regression_range_length}
\mathfrak b_R=1.
\end{equation}
Set
\begin{equation}
\label{eq:app_regression_rho_choice}
\rho_R
=
\frac{\alpha_R}{|\mathcal H_t|}.
\end{equation}
Then
\begin{equation}
\label{eq:app_regression_log_conversion}
\log\frac{2}{\rho_R}
=
\log\frac{2|\mathcal H_t|}{\alpha_R}
=
L_R.
\end{equation}
For this fixed \(h\), Lemma~\ref{lem:app_empirical_bernstein} gives
\begin{equation}
\label{eq:app_regression_concentration_chain}
\begin{aligned}
D_{\mathrm{reg}}(h)-\widehat D_{\mathrm{reg}}(h)
&\le
\sqrt{
\frac{2\widehat V_R(h)L_R}{n_R}
}
+
\frac{7\mathfrak b_RL_R}{3(n_R-1)}
\\
&=
\sqrt{
\frac{2\widehat V_R(h)L_R}{n_R}
}
+
\frac{7L_R}{3(n_R-1)}
\\
&=
\varepsilon_R(h).
\end{aligned}
\end{equation}
Therefore,
\begin{equation}
\label{eq:app_regression_ucb_result}
D_{\mathrm{reg}}(h)
\le
\widehat D_{\mathrm{reg}}(h)+\varepsilon_R(h)
=
\mathrm{UCB}_R(h).
\end{equation}
A union bound over \(h\in\mathcal H_t\) gives
\begin{equation}
\label{eq:app_regression_union_bound}
\mathbb P
\left(
\forall h\in\mathcal H_t:
D_{\mathrm{reg}}(h)
\le
\mathrm{UCB}_R(h)
\right)
\ge
1-\alpha_R.
\end{equation}

Finally consider falsification risk. For a fixed \(h\), apply Lemma~\ref{lem:app_bernoulli_kl} to \(X^F_h\). Let
\begin{equation}
\label{eq:app_failure_u_choice}
u_F
=
\frac{L_F}{n_F}.
\end{equation}
Then
\begin{equation}
\label{eq:app_failure_tail_chain}
\begin{aligned}
\mathbb P
\left(
\mathrm{kl}
\left(
\widehat P_{\mathrm{fail}}(h)\|P_{\mathrm{fail}}(h)
\right)
>
\frac{L_F}{n_F}
\right)
&\le
(n_F+1)\exp(-L_F)
\\
&=
(n_F+1)
\exp
\left(
-\log\frac{(n_F+1)|\mathcal H_t|}{\alpha_F}
\right)
\\
&=
\frac{\alpha_F}{|\mathcal H_t|}.
\end{aligned}
\end{equation}
A union bound over \(h\in\mathcal H_t\) yields
\begin{equation}
\label{eq:app_failure_uniform_kl_event}
\mathbb P
\left(
\forall h\in\mathcal H_t:
\mathrm{kl}
\left(
\widehat P_{\mathrm{fail}}(h)\|P_{\mathrm{fail}}(h)
\right)
\le
\frac{L_F}{n_F}
\right)
\ge
1-\alpha_F.
\end{equation}

On the event in Eq.~\eqref{eq:app_failure_uniform_kl_event}, we prove that \(P_{\mathrm{fail}}(h)\le \mathrm{UCB}_F(h)\) for every \(h\). Fix \(h\). If \(P_{\mathrm{fail}}(h)\le \widehat P_{\mathrm{fail}}(h)\), then
\begin{equation}
\label{eq:app_failure_case_one}
P_{\mathrm{fail}}(h)
\le
\widehat P_{\mathrm{fail}}(h)
\le
\mathrm{UCB}_F(h),
\end{equation}
because \(q=\widehat P_{\mathrm{fail}}(h)\) is feasible in the definition of \(\mathrm{UCB}_F(h)\). If \(P_{\mathrm{fail}}(h)>\widehat P_{\mathrm{fail}}(h)\), then \(q=P_{\mathrm{fail}}(h)\) lies in \([\widehat P_{\mathrm{fail}}(h),1]\), and the KL event gives
\begin{equation}
\label{eq:app_failure_case_two_feasibility}
\mathrm{kl}
\left(
\widehat P_{\mathrm{fail}}(h)\|P_{\mathrm{fail}}(h)
\right)
\le
\frac{L_F}{n_F}.
\end{equation}
Thus \(P_{\mathrm{fail}}(h)\) is feasible in the supremum defining \(\mathrm{UCB}_F(h)\), which implies
\begin{equation}
\label{eq:app_failure_case_two}
P_{\mathrm{fail}}(h)
\le
\mathrm{UCB}_F(h).
\end{equation}
Combining the two cases gives
\begin{equation}
\label{eq:app_failure_ucb_result}
\forall h\in\mathcal H_t:
P_{\mathrm{fail}}(h)
\le
\mathrm{UCB}_F(h).
\end{equation}

Let \(\mathcal E_T\), \(\mathcal E_F\), and \(\mathcal E_R\) denote the target, falsification, and regression confidence events. Combining Eq.~\eqref{eq:app_target_union_bound}, Eq.~\eqref{eq:app_regression_union_bound}, and Eq.~\eqref{eq:app_failure_uniform_kl_event}, we have
\begin{equation}
\label{eq:app_final_union_bound_chain}
\begin{aligned}
\mathbb P(\mathcal E_T\cap\mathcal E_F\cap\mathcal E_R)
&=
1-
\mathbb P(\mathcal E_T^c\cup\mathcal E_F^c\cup\mathcal E_R^c)
\\
&\ge
1
-
\mathbb P(\mathcal E_T^c)
-
\mathbb P(\mathcal E_F^c)
-
\mathbb P(\mathcal E_R^c)
\\
&\ge
1-\alpha_T-\alpha_F-\alpha_R
\\
&=
1-\alpha.
\end{aligned}
\end{equation}
On this event, all three bounds in Eq.~\eqref{eq:app_uniform_confidence_bounds} hold simultaneously for every \(h\in\mathcal H_t\). This proves the lemma.
\end{proof}

\subsection{Proof of Theorem~\ref{thm:commitment_gate_generalization}}
\label{app:proof_commitment_gate_theorem}

\begin{proof}
\label{proof:app_commitment_gate_theorem}
By Lemma~\ref{lem:app_uniform_confidence}, with probability at least \(1-\alpha\), all \(h\in\mathcal H_t\) simultaneously satisfy
\begin{equation}
\label{eq:app_theorem_confidence_event}
G(h)
\ge
\mathrm{LCB}_T(h),
\qquad
P_{\mathrm{fail}}(h)
\le
\mathrm{UCB}_F(h),
\qquad
D_{\mathrm{reg}}(h)
\le
\mathrm{UCB}_R(h).
\end{equation}
This proves Eq.~\eqref{eq:theory_confidence_event}.

Now suppose that \(h\) satisfies the confidence-gated commitment rule:
\begin{equation}
\label{eq:app_confidence_gate_assumption}
\mathrm{LCB}_T(h)\ge \eta,
\qquad
\mathrm{UCB}_F(h)\le \delta,
\qquad
\mathrm{UCB}_R(h)\le \beta.
\end{equation}
For target-transfer gain, Eq.~\eqref{eq:app_theorem_confidence_event} and Eq.~\eqref{eq:app_confidence_gate_assumption} imply
\begin{equation}
\label{eq:app_target_population_chain}
\begin{aligned}
G(h)
&\ge
\mathrm{LCB}_T(h)
\\
&\ge
\eta.
\end{aligned}
\end{equation}
For falsification risk,
\begin{equation}
\label{eq:app_failure_population_chain}
\begin{aligned}
P_{\mathrm{fail}}(h)
&\le
\mathrm{UCB}_F(h)
\\
&\le
\delta.
\end{aligned}
\end{equation}
For regression damage,
\begin{equation}
\label{eq:app_regression_population_chain}
\begin{aligned}
D_{\mathrm{reg}}(h)
&\le
\mathrm{UCB}_R(h)
\\
&\le
\beta.
\end{aligned}
\end{equation}
Combining Eq.~\eqref{eq:app_target_population_chain}, Eq.~\eqref{eq:app_failure_population_chain}, and Eq.~\eqref{eq:app_regression_population_chain}, we obtain
\begin{equation}
\label{eq:app_population_constraints_final}
G(h)\ge\eta,
\qquad
P_{\mathrm{fail}}(h)\le\delta,
\qquad
D_{\mathrm{reg}}(h)\le\beta.
\end{equation}
This proves Eq.~\eqref{eq:theory_population_safety}.
\end{proof}

\subsection{Exact Probe Complexity Statement}
\label{app:exact_probe_complexity}

We now give the exact finite-sample version of Corollary~\ref{cor:probe_complexity_safe_commitment}. Let \(\zeta_T,\zeta_R>0\) be target and regression margins, and let \(0\le\delta_-<\delta\le1\) be empirical and population falsification thresholds.

\begin{corollary}[Exact probe complexity]
\label{cor:app_exact_probe_complexity}
Suppose
\begin{equation}
\label{eq:app_exact_sample_complexity_target}
n_T
\ge
1+
\max
\left\{
\frac{8L_T}{\zeta_T^2},
\frac{28L_T}{3\zeta_T}
\right\},
\end{equation}
\begin{equation}
\label{eq:app_exact_sample_complexity_regression}
n_R
\ge
1+
\max
\left\{
\frac{2L_R}{\zeta_R^2},
\frac{14L_R}{3\zeta_R}
\right\},
\end{equation}
and
\begin{equation}
\label{eq:app_exact_sample_complexity_failure}
n_F
\ge
\frac{L_F}{\mathrm{kl}(\delta_-\|\delta)}.
\end{equation}
Then, with probability at least \(1-\alpha\), any \(h\in\mathcal H_t\) satisfying
\begin{equation}
\label{eq:app_exact_empirical_margin_gate}
\widehat G(h)\ge \eta+\zeta_T,
\qquad
\widehat P_{\mathrm{fail}}(h)\le \delta_-,
\qquad
\widehat D_{\mathrm{reg}}(h)\le \beta-\zeta_R
\end{equation}
also satisfies
\begin{equation}
\label{eq:app_exact_population_constraints}
G(h)\ge\eta,
\qquad
P_{\mathrm{fail}}(h)\le\delta,
\qquad
D_{\mathrm{reg}}(h)\le\beta.
\end{equation}
\end{corollary}

\subsection{Proof of Corollary~\ref{cor:app_exact_probe_complexity}}
\label{app:proof_probe_complexity_corollary}

\begin{proof}
\label{proof:app_probe_complexity_corollary}
We first control the target-transfer confidence radius. Since \(X^T_h\in[-1,1]\), its empirical variance satisfies
\begin{equation}
\label{eq:app_target_variance_worst_case}
\widehat V_T(h)
\le
1.
\end{equation}
Therefore,
\begin{equation}
\label{eq:app_target_slack_worst_case_chain}
\begin{aligned}
\varepsilon_T(h)
&=
\sqrt{
\frac{2\widehat V_T(h)L_T}{n_T}
}
+
\frac{14L_T}{3(n_T-1)}
\\
&\le
\sqrt{
\frac{2L_T}{n_T}
}
+
\frac{14L_T}{3(n_T-1)}.
\end{aligned}
\end{equation}
If
\begin{equation}
\label{eq:app_target_first_sample_condition}
n_T
\ge
\frac{8L_T}{\zeta_T^2},
\end{equation}
then
\begin{equation}
\label{eq:app_target_first_term_bound}
\sqrt{
\frac{2L_T}{n_T}
}
\le
\frac{\zeta_T}{2}.
\end{equation}
If
\begin{equation}
\label{eq:app_target_second_sample_condition}
n_T-1
\ge
\frac{28L_T}{3\zeta_T},
\end{equation}
then
\begin{equation}
\label{eq:app_target_second_term_bound}
\frac{14L_T}{3(n_T-1)}
\le
\frac{\zeta_T}{2}.
\end{equation}
Combining Eq.~\eqref{eq:app_target_slack_worst_case_chain}, Eq.~\eqref{eq:app_target_first_term_bound}, and Eq.~\eqref{eq:app_target_second_term_bound}, we get
\begin{equation}
\label{eq:app_target_slack_final}
\varepsilon_T(h)
\le
\zeta_T.
\end{equation}

We next control the regression confidence radius. Since \(X^R_h\in[0,1]\), its empirical variance satisfies
\begin{equation}
\label{eq:app_regression_variance_worst_case}
\widehat V_R(h)
\le
\frac{1}{4}.
\end{equation}
Thus,
\begin{equation}
\label{eq:app_regression_slack_worst_case_chain}
\begin{aligned}
\varepsilon_R(h)
&=
\sqrt{
\frac{2\widehat V_R(h)L_R}{n_R}
}
+
\frac{7L_R}{3(n_R-1)}
\\
&\le
\sqrt{
\frac{L_R}{2n_R}
}
+
\frac{7L_R}{3(n_R-1)}.
\end{aligned}
\end{equation}
If
\begin{equation}
\label{eq:app_regression_first_sample_condition}
n_R
\ge
\frac{2L_R}{\zeta_R^2},
\end{equation}
then
\begin{equation}
\label{eq:app_regression_first_term_bound}
\sqrt{
\frac{L_R}{2n_R}
}
\le
\frac{\zeta_R}{2}.
\end{equation}
If
\begin{equation}
\label{eq:app_regression_second_sample_condition}
n_R-1
\ge
\frac{14L_R}{3\zeta_R},
\end{equation}
then
\begin{equation}
\label{eq:app_regression_second_term_bound}
\frac{7L_R}{3(n_R-1)}
\le
\frac{\zeta_R}{2}.
\end{equation}
Combining Eq.~\eqref{eq:app_regression_slack_worst_case_chain}, Eq.~\eqref{eq:app_regression_first_term_bound}, and Eq.~\eqref{eq:app_regression_second_term_bound}, we obtain
\begin{equation}
\label{eq:app_regression_slack_final}
\varepsilon_R(h)
\le
\zeta_R.
\end{equation}

We now control the falsification upper confidence bound. Assume \(\widehat P_{\mathrm{fail}}(h)\le\delta_-\). For fixed \(q=\delta\), the function \(p\mapsto \mathrm{kl}(p\|\delta)\) is decreasing on \(p\in[0,\delta]\). Indeed,
\begin{equation}
\label{eq:app_kl_derivative}
\frac{\partial}{\partial p}
\mathrm{kl}(p\|\delta)
=
\log
\frac{
p(1-\delta)
}{
\delta(1-p)
},
\end{equation}
and for \(p<\delta\),
\begin{equation}
\label{eq:app_kl_derivative_negative}
\log
\frac{
p(1-\delta)
}{
\delta(1-p)
}
<
0.
\end{equation}
Therefore,
\begin{equation}
\label{eq:app_kl_empirical_lower_bound}
\mathrm{kl}
\left(
\widehat P_{\mathrm{fail}}(h)\|\delta
\right)
\ge
\mathrm{kl}(\delta_-\|\delta).
\end{equation}
By Eq.~\eqref{eq:app_exact_sample_complexity_failure},
\begin{equation}
\label{eq:app_failure_radius_bound}
\frac{L_F}{n_F}
\le
\mathrm{kl}(\delta_-\|\delta)
\le
\mathrm{kl}
\left(
\widehat P_{\mathrm{fail}}(h)\|\delta
\right).
\end{equation}
For fixed \(p=\widehat P_{\mathrm{fail}}(h)\), the map \(q\mapsto \mathrm{kl}(p\|q)\) is increasing on \(q\in[p,1]\). Hence every \(q>\delta\) satisfies
\begin{equation}
\label{eq:app_failure_q_above_delta}
\mathrm{kl}
\left(
\widehat P_{\mathrm{fail}}(h)\|q
\right)
>
\frac{L_F}{n_F}.
\end{equation}
Thus no \(q>\delta\) is feasible in the supremum defining \(\mathrm{UCB}_F(h)\), and we have
\begin{equation}
\label{eq:app_failure_ucb_delta}
\mathrm{UCB}_F(h)
\le
\delta.
\end{equation}

Now suppose \(h\) satisfies the empirical margin condition in Eq.~\eqref{eq:app_exact_empirical_margin_gate}. Combining Eq.~\eqref{eq:app_target_slack_final}, Eq.~\eqref{eq:app_regression_slack_final}, and Eq.~\eqref{eq:app_failure_ucb_delta}, we get
\begin{equation}
\label{eq:app_margin_implies_confidence_gate}
\mathrm{LCB}_T(h)
\ge
\eta,
\qquad
\mathrm{UCB}_F(h)
\le
\delta,
\qquad
\mathrm{UCB}_R(h)
\le
\beta.
\end{equation}
Applying Theorem~\ref{thm:commitment_gate_generalization} yields
\begin{equation}
\label{eq:app_probe_complexity_population_final}
G(h)\ge\eta,
\qquad
P_{\mathrm{fail}}(h)\le\delta,
\qquad
D_{\mathrm{reg}}(h)\le\beta.
\end{equation}
This proves the corollary.
\end{proof}

\subsection{Remarks on Adaptivity and Scope Refinement}
\label{app:adaptivity_and_refinement}

The proof requires that the final evaluation probes are not reused to construct or refine the hypothesis being certified. This condition prevents adaptive overfitting to the finite falsification set. In practice, this can be implemented in two equivalent ways. The first is sample splitting: use a development split for scope refinement and a fresh evaluation split for final commitment. The second is to define \(\mathcal H_t\) as the enlarged finite set containing all hypotheses that can be produced by proposal and refinement before final evaluation; the same proof then applies with \(|\mathcal H_t|\) replaced by the size of this enlarged set.

The theorem certifies reliability with respect to the probe-generation distributions \(\mathcal D_T(h)\), \(\mathcal D_F(h)\), and \(\mathcal D_R(h)\). If these distributions differ from the deployment distribution, an additional distribution-shift term is needed. For example, if the deployment falsification distribution is within total variation distance \(\rho\) of \(\mathcal D_F(h)\), then the certified falsification risk increases by at most \(\rho\). The commitment gate controls statistical generalization from finite probes to the specified probe distributions, while the quality of the probe generator determines how well those distributions cover deployment-time boundary cases.

\section{Additional Experimental Details}
\label{app:exp_details}

\subsection{Benchmark Details}
\label{app:benchmark_details}

\paragraph{Web and browser-control tasks.}
For WebArena~\citep{zhou2024webarena} and BrowserGym~\citep{chezelles2025browsergym}, we follow the official environment setup and success validators. 
Each task provides a natural-language instruction, an initial browser state, and an environment-specific completion condition. 
We use these validators as the verifier $v_h$ whenever possible. 
For FGSE, target-transfer probes are constructed from tasks sharing the same high-level interaction pattern, while falsification probes are selected from tasks with similar surface forms but different completion conditions, websites, forms, or required constraints. 
Regression probes are drawn from previously solved held-out tasks and are never used to propose the candidate hypothesis.

\paragraph{Tool-use and app-interaction tasks.}
For AppWorld~\citep{trivedi2024appworld}, we evaluate on the official Normal and Challenge splits and report task-goal completion using the benchmark-provided database-state evaluator. 
For $\tau$-bench~\citep{yao2024taubench}, we evaluate both retail and airline domains and use the official database-state comparison and $\mathrm{pass}^k$ reliability protocol. 
In both benchmarks, a hypothesis may modify the agent's tool-use policy, retrieval memory, or instruction-level strategy, but the available tools, API schemas, and domain policies are fixed across all methods. 
Holding tool access fixed isolates the effect of the update mechanism.

\paragraph{Security and adversarial tool-output tasks.}
AgentDojo~\citep{debenedetti2024agentdojo} is used to evaluate whether self-evolution increases susceptibility to prompt injection or malicious tool outputs. 
We report utility success on benign tasks, security success under attacks, and the joint utility-security score. 
For FGSE, falsification probes include decoy tool outputs, conflicting instructions, and previously solved benign cases. 
A candidate update is rejected if it improves utility but increases attack success or violates a high-priority system or user constraint.

\paragraph{Long-term memory tasks.}
For LongMemEval~\citep{wu2024longmemeval} and LoCoMo~\citep{maharana2024locomo}, we evaluate whether committed hypotheses improve memory use without corrupting persistent user or task state. 
We report the official QA accuracy or F1 metrics, temporal reasoning accuracy, knowledge-update accuracy, and abstention accuracy when available. 
We also measure memory pollution rate, defined as the fraction of committed updates that are later contradicted by held-out evidence or cause a regression on archived successful examples.

\paragraph{Code and mathematical reasoning diagnostics.}
We use HumanEval~\citep{chen2021codex}, MBPP~\citep{austin2021program}, and SWE-bench Verified~\citep{chowdhury2024swebenchverified} as verifier-rich coding environments because unit tests provide executable feedback. 
SWE-bench Verified is a human-validated subset of the original SWE-bench~\citep{jimenez2024swebench}, released by OpenAI in collaboration with the SWE-bench authors~\citep{chowdhury2024swebenchverified}.
We report pass@1 on HumanEval and MBPP and resolved rate on SWE-bench Verified. 
For mathematical reasoning, we use GSM8K~\citep{cobbe2021gsm8k} and MATH~\citep{hendrycks2021math}, reporting exact-match solve rate after answer normalization. 
These datasets are used primarily for controlled analysis of update quality because their verifiers allow clean construction of target, boundary, and regression probes.

\subsection{Evaluation Metrics}
\label{app:metrics_details}

Let $\mathcal D$ denote a held-out evaluation set and let $v(x,\tau)\in[0,1]$ be the benchmark-specific verifier score. 
The standard task success rate is
\begin{equation}
\mathrm{SR}
=
\frac{1}{|\mathcal D|}
\sum_{x\in \mathcal D}
v(x,\tau_x).
\end{equation}
For binary task validators, this reduces to the fraction of solved tasks. 
For benchmarks with official non-binary scores, we use the official normalized score.

For each committed or rejected hypothesis $h$, we compute the three FGSE-specific statistics:
\begin{equation}
\widehat G(h)
=
\frac{1}{|\mathcal T(h)|}
\sum_{x\in\mathcal T(h)}
\left[J_{v_h}(\Theta\oplus h,x)-J_{v_h}(\Theta,x)\right],
\end{equation}
\begin{equation}
\widehat P_{\mathrm{fail}}(h)
=
\frac{1}{|\mathcal F(h)|}
\sum_{x\in\mathcal F(h)}
\mathbb I
\left[
J_{v_h}(\Theta\oplus h,x)-J_{v_h}(\Theta,x)<-\epsilon
\;\vee\;
E_h(\Theta\oplus h,x)=0
\right],
\end{equation}
\begin{equation}
\widehat D_{\mathrm{reg}}(h)
=
\frac{1}{|\mathcal R(h)|}
\sum_{x\in\mathcal R(h)}
\left[
J_{v_h}(\Theta,x)-J_{v_h}(\Theta\oplus h,x)
\right]_+.
\end{equation}
We also report:
\begin{itemize}
    \item \textbf{Harmful update rate}: the fraction of committed updates that cause held-out regression larger than $\epsilon$.
    \item \textbf{Accepted-update precision}: the fraction of committed updates that yield positive held-out target-transfer gain without measurable regression.
    \item \textbf{Falsification hit rate}: the fraction of rejected or refined updates for which at least one falsification probe exposes a failure.
    \item \textbf{Post-commit violation rate}: the fraction of future activations where a committed hypothesis violates its expected-effect checker.
    \item \textbf{Cost}: total LLM calls, verifier calls, wall-clock time, and token usage per committed update.
\end{itemize}

\subsection{Fair Comparison Protocol}
\label{app:fair_comparison}

All controlled comparisons follow the same protocol~\citep{yang2026measurementfirst}:
\begin{enumerate}
    \item \textbf{Same backbone}: every method uses the same base LLM and the same ReAct-style executor unless the method intrinsically requires a different executor.
    \item \textbf{Same tools}: all methods receive identical tool schemas, browser actions, API access, and environment observations.
    \item \textbf{Same memory budget}: methods with persistent memory are allowed the same maximum number of stored units and the same retrieval top-$k$.
    \item \textbf{Same update budget}: methods that propose self-updates receive the same number of completed training episodes and the same maximum number of update attempts.
    \item \textbf{No test leakage}: hypotheses, reflections, playbooks, workflows, and memory entries are constructed only from training or development episodes. Test tasks are used only for final evaluation.
    \item \textbf{Same verifier access}: if a verifier is available during development, all methods may use it under the same budget. During final evaluation, verifier outputs are used only for scoring.
\end{enumerate}
For workflow-search or architecture-search methods such as AFlow~\citep{zhang2025aflow}, ADAS~\citep{hu2025adas}, GPTSwarm~\citep{zhuge2024gptswarm}, and MetaAgent~\citep{zhang2025metaagent}, we report a controlled-budget version when code and prompts are available. 
Methods requiring a substantially larger search budget, extra training data, reinforcement learning, or model fine-tuning are reported in a separate native-setting table, outside the controlled FGSE comparisons.

\subsection{Baseline Implementation Details}
\label{app:baseline_details}

\paragraph{Direct and ReAct.}
The direct baseline uses the base LLM without persistent memory or post-episode updates. 
The ReAct baseline~\citep{yao2023react} uses the same action space and tool interface as FGSE but does not modify its persistent state after each episode.

\paragraph{Self-Refine and Reflexion.}
Self-Refine~\citep{madaan2023selfrefine} performs within-episode critique and revision under the same maximum number of refinement calls as FGSE's hypothesis proposal calls. 
Reflexion~\citep{shinn2023reflexion} writes natural-language reflections after failed or low-scoring episodes and retrieves them in future tasks. 
Reflexion commits reflections without requiring structured preconditions, falsification probes, or regression tests.

\paragraph{Voyager-style skill memory.}
The Voyager-style baseline~\citep{wang2023voyager} stores reusable skills or lessons from successful episodes and retrieves them for related future tasks. 
For non-Minecraft benchmarks, we adapt the skill library into a text-based rule or tool-use strategy library while keeping the same retrieval budget as FGSE.

\paragraph{Memory and context-evolution methods.}
A-Mem~\citep{xu2025amem} is implemented as an agentic memory organization baseline that dynamically indexes, links, and updates memories. 
ACE~\citep{zhang2026ace} is implemented as an evolving playbook baseline that incrementally revises persistent context using execution feedback. 
Both methods are evaluated with the same memory capacity and retrieval top-$k$ as FGSE.

\paragraph{Workflow and agent-design optimization methods.}
AFlow~\citep{zhang2025aflow}, ADAS~\citep{hu2025adas}, GPTSwarm~\citep{zhuge2024gptswarm}, and MetaAgent~\citep{zhang2025metaagent} provide comparisons at the agentic-workflow level, complementing update-level memory methods. 
Because these methods may search over a larger design space than FGSE, we distinguish controlled-budget and native-budget results.

\subsection{FGSE Hyperparameters}
\label{app:fgse_hyperparameters}

\begin{table}[t]
\centering
\small
\caption{Default FGSE hyperparameters used in controlled experiments.}
\label{tab:fgse_hyperparams}
\begin{tabular}{lcc}
\toprule
Hyperparameter & Symbol & Default value \\
\midrule
Candidate hypotheses per episode & $B$ & $3$ \\
Target-transfer probes & $|\mathcal T(h)|$ & $8$ \\
Falsification probes & $|\mathcal F(h)|$ & $16$ \\
Regression probes & $|\mathcal R(h)|$ & $16$ \\
Minimum target gain & $\eta$ & $0.03$ \\
Maximum falsification rate & $\delta$ & $0.05$ \\
Maximum regression damage & $\beta$ & $0.01$ \\
Tolerated verifier-score drop & $\epsilon$ & $0$ \\
Precondition activation threshold & $\gamma_h$ & $0.5$ \\
Intervention strength & $\lambda_h$ & $0.8$ \\
Retrieved committed hypotheses & $k$ & $5$ \\
Post-commit violation threshold & $\rho_{\max}$ & $0.10$ \\
\bottomrule
\end{tabular}
\end{table}

All hyperparameters are selected on development tasks and then fixed for test evaluation. 

\subsection{Ablation Studies}
\label{app:ablations}

We conduct the following ablations:
\begin{itemize}
    \item \textbf{No falsification}: commit every proposed hypothesis that improves the source episode.
    \item \textbf{Random probes}: replace targeted falsification probes with randomly sampled held-out tasks.
    \item \textbf{LLM-judge gate}: replace executable or environment verifiers with an LLM-only judgment.
    \item \textbf{No regression archive}: remove $\mathcal R(h)$ from the commitment rule.
    \item \textbf{No scope narrowing}: reject all hypotheses that fail falsification without allowing precondition refinement.
    \item \textbf{Free-form memory}: store unstructured natural-language lessons instead of structured $(c_h,b_h,e_h,v_h)$ hypotheses.
    \item \textbf{Probe-budget sweep}: vary the number of falsification probes to measure the trade-off between update safety and computation.
\end{itemize}
These ablations isolate whether FGSE's gains come from verifier grounding, targeted counterexample construction, regression protection, structured update representation, or scope refinement.

\subsection{Compute Resources}
\label{app:compute_resources}

FGSE uses frozen language models without training or fine-tuning. All reported
experiments access frozen LLM backbones through provider-hosted APIs, with no
author-side GPU use for model training or inference. The provider-side
serving compute for GPT-4o, GPT-4o-mini, and DeepSeek-V3.1 is not observable to
us and is not included in the reported compute. Our local compute consists of
agent orchestration, environment simulation, browser/app execution, sandboxed
code execution, probe construction, verifier calls, logging, and result
aggregation.

Unless otherwise stated, experiments were orchestrated on CPU workers with
8 vCPUs and 32GB memory per worker. For code-execution and repository-level
diagnostics such as SWE-bench Verified, we used sandbox workers with 16 vCPUs
and 64GB memory. We used up to 32 workers concurrently. We report compute in
terms of LLM calls, verifier calls, parallel wall-clock time, and aggregate
client-side CPU usage. One LLM call denotes one model completion request,
including action generation, reflection, hypothesis proposal, scope refinement,
and LLM-judge calls when used by a baseline. One verifier call denotes one
execution of an official benchmark validator, unit test, database-state checker,
security checker, or exact-match evaluator.

\begin{table}[t]
\centering
\small
\caption{
\textbf{Approximate compute used by the reported experiments.}
LLM calls include agent rollouts, hypothesis proposal/refinement, and
baseline-specific reflection or judging calls. Verifier calls include official
success validators, database-state checkers, unit tests, security validators,
and exact-match evaluators. CPU-hours denote client-side orchestration and
environment/verifier execution only. Provider-side LLM serving compute is not
included because all LLM backbones are accessed through hosted APIs.
}
\label{tab:compute_resources}
\resizebox{\linewidth}{!}{
\begin{tabular}{lrrrr}
\toprule
\textbf{Experiment group}
& \textbf{LLM calls}
& \textbf{Verifier calls}
& \textbf{Wall-clock time}
& \textbf{Client CPU-hours} \\
\midrule
Web/browser-control tasks
& $\approx 7.8{\times}10^4$
& $\approx 1.1{\times}10^5$
& $\approx 39$ h
& $\approx 1.1{\times}10^3$ \\
App/API-interaction tasks
& $\approx 1.46{\times}10^5$
& $\approx 2.18{\times}10^5$
& $\approx 32$ h
& $\approx 9.0{\times}10^2$ \\
$\tau$-bench tool-user tasks
& $\approx 8.4{\times}10^4$
& $\approx 1.24{\times}10^5$
& $\approx 28$ h
& $\approx 7.2{\times}10^2$ \\
AgentDojo security diagnostics
& $\approx 3.6{\times}10^4$
& $\approx 6.8{\times}10^4$
& $\approx 12$ h
& $\approx 3.0{\times}10^2$ \\
Long-term memory benchmarks
& $\approx 6.4{\times}10^4$
& $\approx 8.7{\times}10^4$
& $\approx 16$ h
& $\approx 4.2{\times}10^2$ \\
Code/math diagnostics
& $\approx 5.2{\times}10^4$
& $\approx 9.6{\times}10^4$
& $\approx 21$ h
& $\approx 7.6{\times}10^2$ \\
Ablations and sensitivity analyses
& $\approx 9.5{\times}10^4$
& $\approx 1.38{\times}10^5$
& $\approx 27$ h
& $\approx 7.4{\times}10^2$ \\
\midrule
\textbf{Total reported experiments}
& $\approx 5.55{\times}10^5$
& $\approx 8.41{\times}10^5$
& $\approx 175$ h
& $\approx 4.94{\times}10^3$ \\
\bottomrule
\end{tabular}
}
\end{table}

The full project also included preliminary diagnostics, pilot prompt variants,
failed verifier configurations, and exploratory ablations that are not included
in Table~\ref{tab:compute_resources}. These exploratory runs used approximately
$1.6\times$ the compute of the final reported experiments. In total, the project
used approximately $8.9{\times}10^5$ LLM calls, $1.35{\times}10^6$ verifier
calls, and $7.9{\times}10^3$ client CPU-hours. Most additional compute came from
early probe-construction variants and unsuccessful target-only or LLM-judge
gating configurations.

\section{Additional Experimental Results}
\label{app:additional_results}
\subsection{Additional Reliability and Ablation Results}
\label{app:reliability_ablation}

\paragraph{Memory and update reliability.}
\label{app.meo}
Table~\ref{tab:main_memory_reliability_results} further evaluates long-term memory performance and controlled self-evolution reliability. 
FGSE obtains the strongest memory performance while also reducing unsafe update commitment, falsification failure, and regression damage. 
These results support improving future target performance while preserving previously solved behaviors and rejecting overgeneralized updates.

\begin{table*}[t]
\centering
\scriptsize
\caption{
\textbf{Long-term memory and controlled update-reliability comparison.}
The Offline Reading result on LongMemEval is an oracle-style upper bound and is not ranked.
Gray $0.0$ entries indicate methods without persistent self-update, for which update-reliability metrics are not ranked.
$\widehat G$ measures target-transfer gain; $\widehat P_{\mathrm{fail}}$ measures falsification failure rate; $\widehat D_{\mathrm{reg}}$ measures regression damage.
Lower is better for the last four columns among self-updating methods.
Best and second-best ranked scores in each column are shown in \textbf{bold} and \underline{underlined}, respectively.
}
\label{tab:main_memory_reliability_results}
\resizebox{\textwidth}{!}{
\begin{tabular}{llccccccc}
\toprule
\textbf{Method}
& \textbf{Backbone / Setting}
& \makecell{\textbf{LongMemEval}\\\textbf{Acc.} $\uparrow$}
& \makecell{\textbf{LoCoMo}\\\textbf{Avg. R-L} $\uparrow$}
& $\widehat G$ (\%) $\uparrow$
& $\widehat P_{\mathrm{fail}}$ (\%) $\downarrow$
& $\widehat D_{\mathrm{reg}}$ (\%) $\downarrow$
& \makecell{\textbf{Harmful}\\\textbf{Updates} (\%) $\downarrow$}
& \makecell{\textbf{Post-commit}\\\textbf{Violations} (\%) $\downarrow$} \\
\midrule

\grouprowmem{Web / browser-control agents: methods primarily designed or evaluated for browser navigation and web-interaction self-improvement.}

GPT-4 Agent~\citep{zhou2024webarena}
& GPT-4
& 0.552
& 22.84
& \inactive{0.0}
& \inactive{0.0}
& \inactive{0.0}
& \inactive{0.0}
& \inactive{0.0} \\

AutoEval~\citep{pan2024autoeval}
& GPT-4
& 0.548
& 23.11
& 2.4
& 19.7
& 5.3
& 13.2
& 11.7 \\

BrowserGym~\citep{chezelles2025browsergym}
& GPT-4o
& 0.560
& 24.02
& 2.8
& 18.9
& 5.1
& 12.8
& 11.1 \\

CER$_{\mathrm{online}}$~\citep{liu2025cer}
& GPT-4o
& 0.591
& 26.37
& 8.6
& 17.4
& 4.9
& 12.1
& 10.8 \\

CER$_{\mathrm{offline}}$~\citep{liu2025cer}
& GPT-4o
& 0.598
& 26.91
& 8.8
& 17.1
& 4.8
& 11.9
& 10.5 \\

CER$_{\mathrm{hybrid}}$~\citep{liu2025cer}
& GPT-4o
& 0.607
& 27.58
& 9.1
& 16.8
& 4.7
& 11.5
& 10.2 \\

\midrule
\grouprowmem{App / API-interaction agents: methods primarily evaluated on app-level tool use, database-state goals, and API execution.}

IPFunCall~\citep{trivedi2024appworld}
& GPT-4o
& 0.507
& 18.24
& 1.9
& 27.6
& 7.0
& 20.1
& 17.5 \\

FullCodeRefl~\citep{trivedi2024appworld}
& GPT-4o
& 0.512
& 18.69
& 3.2
& 24.9
& 6.5
& 18.4
& 15.9 \\

PlanExec~\citep{trivedi2024appworld}
& GPT-4o
& 0.518
& 20.15
& 3.8
& 23.5
& 6.2
& 17.9
& 15.2 \\

ReAct~\citep{yao2023react,trivedi2024appworld}
& GPT-4o
& 0.560
& 22.46
& \inactive{0.0}
& \inactive{0.0}
& \inactive{0.0}
& \inactive{0.0}
& \inactive{0.0} \\

ReAct + ICL~\citep{zhang2026ace}
& DeepSeek-V3.1
& 0.572
& 23.46
& 6.4
& 18.6
& 4.9
& 13.1
& 11.4 \\

ReAct + GEPA~\citep{agrawal2026gepa,zhang2026ace}
& DeepSeek-V3.1
& 0.575
& 23.91
& 7.2
& 17.9
& 4.7
& 12.4
& 10.9 \\

ReAct + DC~\citep{suzgun2025dc}
& DeepSeek-V3.1
& 0.590
& 26.42
& 9.4
& 14.8
& 4.4
& 10.9
& 9.6 \\

ReAct + ACE$_{\mathrm{offline}}$~\citep{zhang2026ace}
& DeepSeek-V3.1
& 0.661
& 31.86
& 10.3
& 12.4
& 3.6
& 8.4
& 7.3 \\

ReAct + ACE$_{\mathrm{online}}$~\citep{zhang2026ace}
& DeepSeek-V3.1
& 0.672
& 32.91
& 10.8
& 11.6
& 3.3
& 7.9
& 6.8 \\

\midrule
\grouprowmem{Tool-user interaction agents: methods centered on multi-turn user simulation, tool calling, and domain-policy execution.}

Tool-Calling~\citep{yao2024taubench}
& GPT-4o
& 0.557
& 22.15
& 2.6
& 22.8
& 6.1
& 16.9
& 14.6 \\

Tool-Calling~\citep{yao2024taubench}
& Claude-3.5-Sonnet
& 0.581
& 24.63
& 3.1
& 21.4
& 5.8
& 15.7
& 13.9 \\

Act~\citep{yao2024taubench}
& GPT-4o
& 0.538
& 21.37
& 2.3
& 23.6
& 6.4
& 17.6
& 15.1 \\

ReAct~\citep{yao2023react,yao2024taubench}
& GPT-4o
& 0.545
& 21.80
& \inactive{0.0}
& \inactive{0.0}
& \inactive{0.0}
& \inactive{0.0}
& \inactive{0.0} \\

\midrule
\grouprowmem{Long-term memory agents: methods primarily designed for persistent memory, conversational memory, or memory-augmented retrieval.}

Offline Reading~\citep{wu2024longmemeval}
& GPT-4o oracle
& \inactive{0.918}
& 30.15
& \inactive{0.0}
& \inactive{0.0}
& \inactive{0.0}
& \inactive{0.0}
& \inactive{0.0} \\

Coze~\citep{wu2024longmemeval}
& GPT-4o
& 0.330
& 18.90
& 4.1
& 26.8
& 6.9
& 20.7
& 17.2 \\

ChatGPT~\citep{wu2024longmemeval}
& GPT-4o
& 0.577
& 25.32
& \inactive{0.0}
& \inactive{0.0}
& \inactive{0.0}
& \inactive{0.0}
& \inactive{0.0} \\

GPT-4o + Chain-of-Note~\citep{wu2024longmemeval}
& long-context
& 0.640
& 29.84
& 5.6
& 19.9
& 5.5
& 14.8
& 12.6 \\

ChatGPT~\citep{wu2024longmemeval}
& GPT-4o-mini
& 0.711
& 27.66
& \inactive{0.0}
& \inactive{0.0}
& \inactive{0.0}
& \inactive{0.0}
& \inactive{0.0} \\

ReadAgent~\citep{xu2025amem}
& GPT-4o-mini
& 0.612
& 9.61
& 4.8
& 24.2
& 6.3
& 18.8
& 15.7 \\

MemoryBank~\citep{xu2025amem}
& GPT-4o-mini
& 0.598
& 6.96
& 4.2
& 25.7
& 6.6
& 19.6
& 16.4 \\

MemGPT~\citep{xu2025amem}
& GPT-4o-mini
& 0.630
& 29.19
& 6.7
& 23.1
& 6.0
& 17.3
& 14.9 \\

LoCoMo~\citep{maharana2024locomo,xu2025amem}
& GPT-4o-mini
& 0.651
& 32.68
& 7.3
& 20.5
& 5.4
& 15.5
& 13.4 \\

A-Mem~\citep{xu2025amem}
& GPT-4o-mini
& 0.684
& 35.49
& 9.9
& 13.2
& 3.8
& 8.7
& 7.6 \\

\midrule
\grouprowmem{FGSE variants: the same falsification-gated update mechanism instantiated with different base LLM backbones.}

\rowcolor{oursblue}
FGSE (Ours)
& GPT-4o-mini
& 0.721
& 36.41
& \second{12.5}
& 6.2
& 1.8
& 4.5
& 3.9 \\

\rowcolor{oursblue}
FGSE (Ours)
& GPT-4o
& \second{0.736}
& \second{36.98}
& 12.1
& \second{5.5}
& \second{1.5}
& \second{4.0}
& \second{3.5} \\

\rowcolor{oursblue}
FGSE (Ours)
& DeepSeek-V3.1
& \best{0.748}
& \best{37.62}
& \best{12.9}
& \best{5.0}
& \best{1.3}
& \best{3.6}
& \best{3.1} \\

\bottomrule
\end{tabular}
}
\vspace{-1mm}
\end{table*}

\paragraph{Additional ablation results}
Table\ref{tab:ablation_update_reliability} presents the results of additional ablation experiments.

\begin{table*}[t]
\centering
\scriptsize
\caption{
\textbf{Single-factor ablation on update reliability.}
$\widehat G$ measures target-transfer gain; $\widehat P_{\mathrm{fail}}$ measures falsification failure rate; $\widehat D_{\mathrm{reg}}$ measures regression damage.
For $\widehat G$, higher is better; for the remaining metrics, lower is better.
Blue numbers in parentheses denote the absolute degradation compared with the full FGSE model.
}
\label{tab:ablation_update_reliability}
\resizebox{\textwidth}{!}{
\begin{tabular}{lccccc}
\toprule
\textbf{Variant}
& $\widehat G$ (\%) $\uparrow$
& $\widehat P_{\mathrm{fail}}$ (\%) $\downarrow$
& $\widehat D_{\mathrm{reg}}$ (\%) $\downarrow$
& \makecell{\textbf{Harmful}\\\textbf{Updates} (\%) $\downarrow$}
& \makecell{\textbf{Post-commit}\\\textbf{Violations} (\%) $\downarrow$} \\
\midrule

\rowcolor{oursblue}
Full FGSE
& \best{12.9}
& \best{5.0}
& \best{1.3}
& \best{3.6}
& \best{3.1} \\

\quad w/o structured hypothesis
& 9.2\abldrop{-3.7}
& 16.7\abldrop{+11.7}
& 4.8\abldrop{+3.5}
& 11.3\abldrop{+7.7}
& 10.1\abldrop{+7.0} \\

\quad w/o target-transfer probes
& 9.8\abldrop{-3.1}
& 11.9\abldrop{+6.9}
& 3.1\abldrop{+1.8}
& 8.2\abldrop{+4.6}
& 7.0\abldrop{+3.9} \\

\quad w/o falsification probes
& 11.7\abldrop{-1.2}
& 18.6\abldrop{+13.6}
& 4.1\abldrop{+2.8}
& 10.6\abldrop{+7.0}
& 9.4\abldrop{+6.3} \\

\quad w/o regression archive
& 12.2\abldrop{-0.7}
& 8.9\abldrop{+3.9}
& 5.2\abldrop{+3.9}
& 8.1\abldrop{+4.5}
& 7.4\abldrop{+4.3} \\

\quad w/o scope refinement
& 10.6\abldrop{-2.3}
& \second{6.4}\abldrop{+1.4}
& \second{1.8}\abldrop{+0.5}
& \second{4.4}\abldrop{+0.8}
& \second{3.9}\abldrop{+0.8} \\

\quad LLM-judge gate
& 10.1\abldrop{-2.8}
& 13.8\abldrop{+8.8}
& 3.9\abldrop{+2.6}
& 9.4\abldrop{+5.8}
& 8.5\abldrop{+5.4} \\

\quad w/o post-commit monitoring
& \second{12.5}\abldrop{-0.4}
& 7.4\abldrop{+2.4}
& 2.6\abldrop{+1.3}
& 6.5\abldrop{+2.9}
& 10.6\abldrop{+7.5} \\

\bottomrule
\end{tabular}
}
\vspace{-1mm}
\end{table*}

\subsection{Update-Selection and Falsification Analysis}
\label{app:update_selection_analysis}

\paragraph{Plasticity--safety Pareto frontier}
\label{sec:pareto_plasticity_safety}

Effective self-evolution combines useful agent improvements with protection against harmful or regressive memory updates. 
We examine the \emph{plasticity--safety Pareto frontier} for update selection and compare FGSE's position with alternative strategies.

\begin{figure*}[t]
    \centering
    \includegraphics[width=0.98\textwidth]{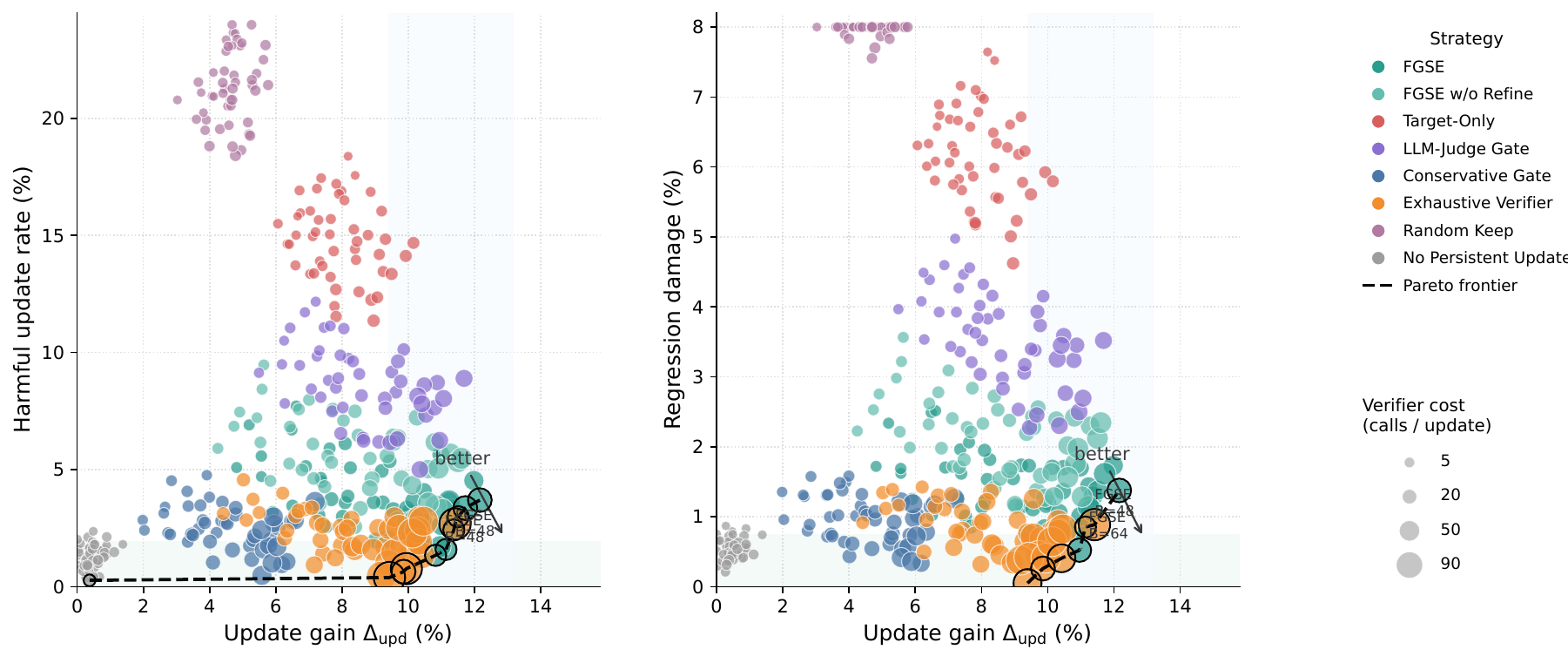}
    \caption{
    \textbf{Plasticity--safety Pareto analysis of update-selection strategies.}
    Each point is one configuration obtained by sweeping verifier budgets and acceptance thresholds.
    The $x$-axis is update gain $\Delta_{\mathrm{upd}}$, and the $y$-axis is either harmful update rate $H$ (left) or regression damage $R$ (right). 
    Point size indicates verifier cost $C$ (average verifier calls per candidate update), and color denotes the update-selection strategy.
    The black dashed line shows the empirical Pareto frontier, where larger gain and smaller harm are jointly preferred.
    FGSE occupies the most favorable lower-right region over a wide range of medium verifier budgets, showing that it improves plasticity without incurring the safety cost typically associated with aggressive update policies.
    }
    \label{fig:pareto_plasticity_safety}
\end{figure*}

We sweep a broad range of update-selection configurations and compare representative strategies, including \textsc{FGSE}, \textsc{FGSE w/o Refine}, \textsc{Target-Only}, \textsc{LLM-Judge Gate}, \textsc{Conservative Gate}, \textsc{Exhaustive Verifier}, \textsc{Random Keep}, and \textsc{No Persistent Update}. 
For each configuration $c$, we measure: (i) \emph{update gain} $\Delta_{\mathrm{upd}}(c)$, defined as the average improvement after applying the selected update; (ii) \emph{harmful update rate} $H(c)$, i.e., the fraction of committed updates that later induce safety violations or incorrect behavior on held-out probes; (iii) \emph{regression damage} $R(c)$, i.e., the average degradation on archived successful behaviors; and (iv) \emph{verifier cost} $C(c)$, measured as the average number of verifier calls per candidate update.
Figure~\ref{fig:pareto_plasticity_safety} visualizes the resulting trade-off. 
The left panel plots $\Delta_{\mathrm{upd}}$ against $H$, while the right panel plots $\Delta_{\mathrm{upd}}$ against $R$. 
Each point corresponds to one configuration, point size encodes verifier cost, and the black dashed curve marks the empirical Pareto frontier.

Figure~\ref{fig:pareto_plasticity_safety} reveals a clear plasticity--safety frontier.
First, \textsc{FGSE} traces the dominant lower-right boundary in both panels over a broad range of medium-cost configurations, achieving substantially larger update gain at similar or lower harm than competing strategies. 
In the harmful-update panel, frontier configurations of \textsc{FGSE} achieve roughly $9\%$--$12\%$ update gain while keeping the harmful update rate within $2.5\%$--$4.5\%$, typically with only $20$--$40$ verifier calls per candidate update. 
At comparable gain levels, \textsc{Target-Only} and \textsc{LLM-Judge Gate} incur much larger harmful update rates, often exceeding $8\%$--$15\%$, indicating that target improvement alone is an unreliable proxy for safe transfer.

Second, a purely conservative policy is also suboptimal.
\textsc{Conservative Gate} combines low harm with much smaller update gain in the lower-left region.
This shows that effective self-improvement also requires sufficient plasticity to capture reusable behavioral rules.
Conversely, \textsc{Exhaustive Verifier} pushes harm slightly lower than \textsc{FGSE} in a few settings, but does so at much higher verifier cost and with little or no gain advantage, indicating that aggressive verification yields diminishing returns.

Third, the comparison between \textsc{FGSE} and \textsc{FGSE w/o Refine} highlights the functional role of scope refinement.
Without refinement, many configurations move upward in both panels: they retain some gain but suffer higher harmful update rates and larger regression damage.
This suggests that a substantial fraction of risky updates retain utility and become safe only after narrowing their applicability conditions.
Overall, the Pareto analysis supports FGSE's better trade-off between \emph{plasticity} and \emph{safety}, placing it in a more favorable region of the update-selection design space.

\paragraph{Targeted probes versus random probes.}
\label{sec:targeted_vs_random_probes}

FGSE uses \emph{targeted} falsification: the verifier constructs probes likely to expose scope overreach, missing preconditions, archived regressions, and decoy-induced shortcut behavior. 
We compare targeted probe scheduling with budget-matched baselines to evaluate its efficiency in discovering harmful updates.

Let $\mathcal{H}_{\mathrm{bad}}$ denote the set of candidate updates that are in fact unsafe or regressive under held-out evaluation. 
Given a probe budget $K$, we define the \emph{bad-update detection rate} as
\begin{equation}
D(K)
=
\frac{1}{|\mathcal{H}_{\mathrm{bad}}|}
\sum_{h \in \mathcal{H}_{\mathrm{bad}}}
\mathbb{1}\!\left[\tau_h \le K\right],
\end{equation}
where $\tau_h$ is the number of probes required to expose update $h$ under a given scheduling policy. 
Thus, $D(K)$ measures the fraction of unsafe updates discovered within the first $K$ probes.

To account for the severity of detected updates, we also report \emph{harmful update avoided},
\begin{equation}
A(K)
=
\frac{
\sum_{h \in \mathcal{H}_{\mathrm{bad}}}
w_h \, \mathbb{1}\!\left[\tau_h \le K\right]
}{
\sum_{h \in \mathcal{H}_{\mathrm{bad}}}
w_h
},
\end{equation}
where $w_h$ is the downstream harm weight of update $h$, estimated from its induced violation severity and regression impact. 
Hence, $A(K)$ measures how much total harmful mass would be prevented if all detected bad updates were rejected before commitment.

\begin{figure*}[t]
    \centering
    \includegraphics[width=0.98\textwidth]{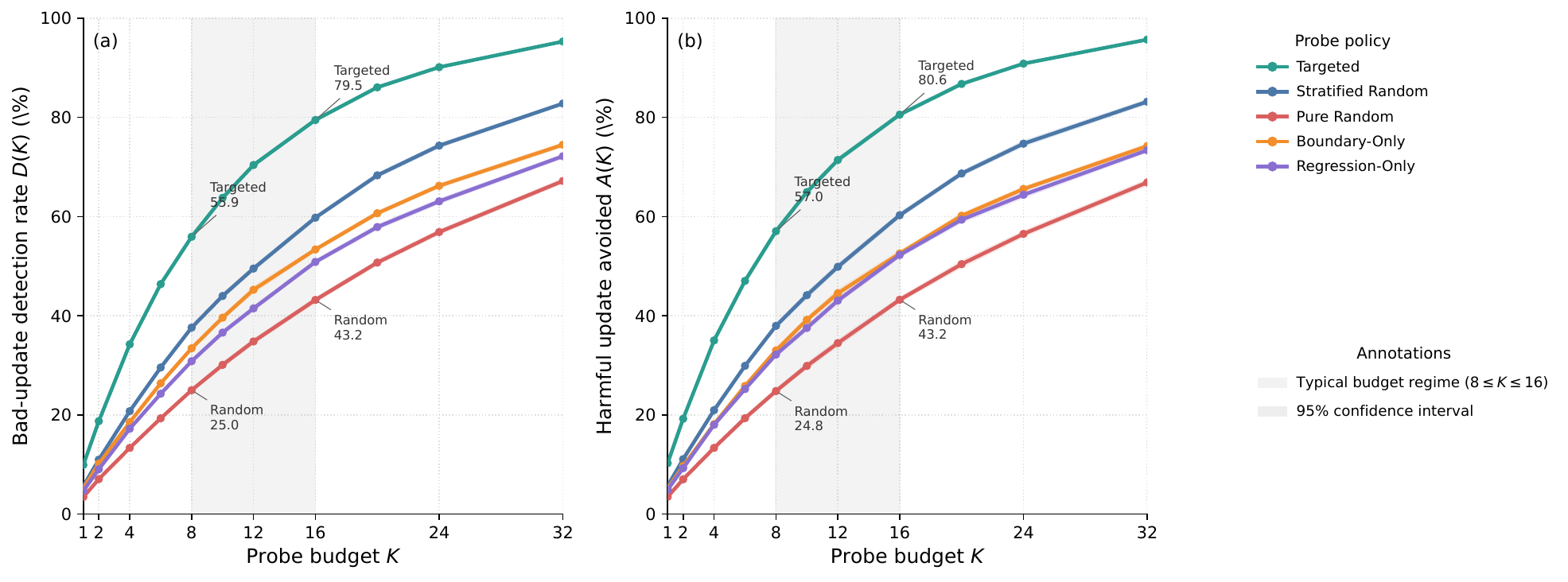}
    \caption{
    \textbf{Falsification discovery curves under different probe-selection policies.}
    The $x$-axis is the probe budget $K$. 
    The left panel shows bad-update detection rate $D(K)$, and the right panel shows harmful update avoided $A(K)$. 
    Solid lines denote the mean across repeated trials, and shaded bands denote $95\%$ confidence intervals. 
    The gray vertical band marks the practically relevant low-budget regime ($8 \le K \le 16$). 
    Targeted probing consistently dominates random baselines, especially in the low-budget regime, showing that hypothesis-aware falsification is substantially more information-efficient than unguided random testing.
    }
    \label{fig:falsification_discovery_curve}
\end{figure*}

We compare five probe-selection policies under the same verifier budget: 
\textsc{Targeted} (the full FGSE probe scheduler), 
\textsc{Stratified Random} (random selection balanced across probe sources), 
\textsc{Pure Random} (uniform sampling from the entire probe pool), 
\textsc{Boundary-Only} (target only applicability-boundary probes), and 
\textsc{Regression-Only} (target only archived-regression probes). 
Figure~\ref{fig:falsification_discovery_curve} plots the resulting falsification discovery curves.

Figure~\ref{fig:falsification_discovery_curve} shows that targeted probing is markedly more informative than random testing at every budget. 
The gap is largest in the practically relevant low-budget regime, where verifier calls are scarce. 
At around $K=8$, the \textsc{Targeted} policy already detects roughly twice as many bad updates as \textsc{Pure Random}, and the improvement is even larger in terms of harmful update avoided. 
This indicates that targeted probing discovers \emph{more} harmful updates early and prioritizes those with larger downstream damage.

A second observation is that \textsc{Boundary-Only} and \textsc{Regression-Only} each improve over pure random sampling in the regimes where their probe families are most relevant, but both eventually saturate below the full targeted strategy. 
This suggests that harmful updates are heterogeneous: some arise from overly broad applicability conditions, whereas others only become visible through archived-regression or decoy-style counterfactual tests. 
Therefore, a single probe family is insufficient; the gain of FGSE comes from selecting probes conditionally on the specific hypothesis under test.

Finally, the larger gap in $A(K)$ than in $D(K)$ is itself informative. 
It shows that targeted probing tends to surface the \emph{most dangerous} updates first while increasing discovery count. 
This property is crucial for practical self-evolution, where only a small verifier budget is available and preventing a few highly harmful commitments is more important than marginally increasing average detection at very large budgets.

\paragraph{Stability under stochastic rollouts.}
\label{sec:stochastic_rollout_stability}

We evaluate each candidate hypothesis under multiple stochastic rollouts to assess the reproducibility of its benefit across seeds.

For a hypothesis $h$ and evaluation set $\mathcal{X}_h$, we define its seed-specific update effect as
\begin{equation}
\Delta_h^{(s)}
=
\frac{1}{|\mathcal{X}_h|}
\sum_{x \in \mathcal{X}_h}
\Big(
R(\pi \oplus h; x, s)
-
R(\pi; x, s)
\Big),
\end{equation}
where $s$ is the rollout seed controlling stochastic decoding, action tie-breaking, and environment perturbations.
Thus, $\Delta_h^{(s)}$ measures the marginal effect of applying hypothesis $h$ under seed $s$.

To quantify stability, we further compute the mean seed effect
\begin{equation}
\bar{\Delta}_h
=
\frac{1}{S}
\sum_{s=1}^{S} \Delta_h^{(s)},
\end{equation}
and the sign-consistency score
\begin{equation}
C(h)
=
\frac{1}{S}
\sum_{s=1}^{S}
\mathbb{1}\!\left[
\operatorname{sign}\!\big(\Delta_h^{(s)}\big)
=
\operatorname{sign}\!\big(\bar{\Delta}_h\big)
\right],
\end{equation}
which measures the fraction of seeds whose update effect agrees in sign with the mean effect. 
A stable and genuinely useful update should exhibit both a positive $\bar{\Delta}_h$ and high sign consistency.

\begin{figure*}[t]
    \centering
    \includegraphics[width=0.98\textwidth]{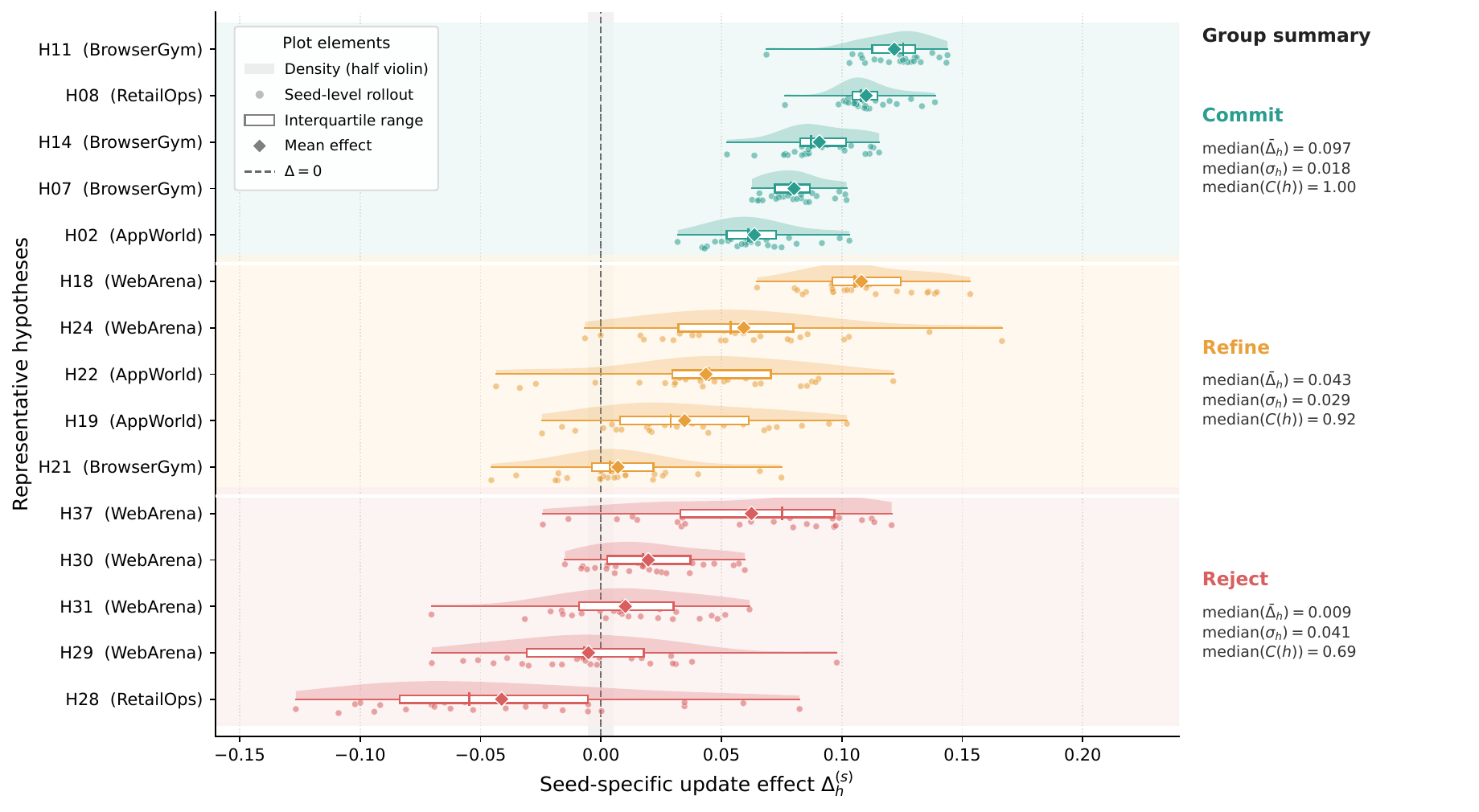}
    \caption{
    \textbf{Stability of hypothesis effects under stochastic rollouts.}
    Each row corresponds to one hypothesis and shows the distribution of seed-specific update effects $\Delta_h^{(s)}$ across repeated rollouts. 
    Rows are grouped by the final FGSE decision. 
    Dots denote seed-level outcomes, the shaded half-violin shows the empirical density, the white box shows the interquartile range, and the diamond marks the mean effect. 
    The right-side annotation reports the sign-consistency score $C(h)$. 
    Committed hypotheses remain consistently positive across seeds, refined hypotheses are still mostly positive but exhibit larger variance, while rejected hypotheses are centered near zero or change sign frequently.
    }
    \label{fig:stochastic_rollout_raincloud}
\end{figure*}

Figure~\ref{fig:stochastic_rollout_raincloud} visualizes the seed-level effect distributions for representative hypotheses from the \textsc{Commit}, \textsc{Refine}, and \textsc{Reject} groups. 
Each row shows the distribution of $\Delta_h^{(s)}$ over seeds for one hypothesis, using a raincloud plot. 
This visualization shows the robustness of observed gains across stochastic rollouts.

Figure~\ref{fig:stochastic_rollout_raincloud} shows a clear separation between update classes.
Committed hypotheses are tightly concentrated on the positive side, with small seed variance and high sign consistency.
In our diagnostic run, the median committed hypothesis has $\bar{\Delta}_h \approx 0.10$ and median sign consistency above $0.9$, indicating that its positive effect persists across seeds.
Refined hypotheses remain beneficial on average, but their distributions are wider and more sensitive to rollout randomness, which is consistent with the interpretation that they capture partially useful rules whose scope still requires tightening.
In contrast, rejected hypotheses are typically centered near zero or slightly negative, and many of them flip sign across seeds, showing that their apparent gains are unstable and largely attributable to stochastic variation.

The effects of accepted updates persist across seeds. Hypothesis identity dominates rollout randomness as a source of variation. FGSE preferentially commits hypotheses whose benefit persists across stochastic executions.

\paragraph{Representative Commit--Refine--Reject Cases}
\label{app:triage_examples}

We inspect three representative hypotheses from the development audit archive. Each case is evaluated
under the default setting used in the main experiments: \(|T(h)|=8\), \(|F(h)|=16\),
\(|R(h)|=16\), \(\eta=0.03\), \(\delta=0.05\), and \(\beta=0.01\). We report the
target-transfer gain \(\widehat{G}(h)\), falsification failure rate
\(\widehat{P}_{\mathrm{fail}}(h)\), regression damage \(\widehat{D}_{\mathrm{reg}}(h)\),
and the resulting gate decision. These cases illustrate how FGSE distinguishes transferable rules, repairable
overgeneralizations, and unsafe shortcuts.

\begin{table}[t]
\centering
\setlength{\tabcolsep}{3.0pt}
\renewcommand{\arraystretch}{1.18}
\caption{
Representative FGSE hypothesis triage cases under the default gate.
All three candidates improve the triggering or near-transfer cases. Based on
falsification and regression evidence, FGSE commits high-margin transferable rules,
refines useful but overly broad rules, and rejects high-risk shortcuts despite
positive target gain.
}
\label{tab:triage_examples}
\resizebox{\textwidth}{!}{%
\begin{tabular}{p{1.25cm}p{2.05cm}p{4.55cm}p{3.45cm}p{3.55cm}}
\toprule
Decision
& Domain / Trigger
& Candidate hypothesis
& Gate statistics
& Interpretation \\
\midrule

\textsc{Commit}
&
\(\tau\)-Airline /
round-trip booking
&
\textbf{Scope:} the user asks to book a round-trip flight and explicitly provides
origin, destination, outbound date, and return date.

\textbf{Behavior:} select the round-trip mode before search, fill both date fields,
and verify that both flight legs are present before final booking.

\textbf{Expected effect:} the final booking record contains
\texttt{trip\_type=round\_trip} and valid outbound and return segments.
&
\(\widehat{G}(h)=12.1\%\)

\(\widehat{P}_{\mathrm{fail}}(h)=0/16=0.0\%\)

\(\widehat{D}_{\mathrm{reg}}(h)=0.6\%\)

Post-commit violation: \(1.1\%\)
&
The hypothesis passes all three default thresholds with margin:
\(\widehat{G}(h)>\eta\),
\(\widehat{P}_{\mathrm{fail}}(h)<\delta\), and
\(\widehat{D}_{\mathrm{reg}}(h)<\beta\).
It captures a reusable tool-order rule without harming boundary or archived cases.
\\
\midrule

\textsc{Refine}
&
AppWorld /
calendar rescheduling
&
\textbf{Initial scope:} any request mentioning ``reschedule''.

\textbf{Behavior:} edit the existing calendar event rather than creating a new one.

\textbf{Failure mode:} the rule also fires on tasks that ask to create a second
event, cancel an event, or modify only a reminder.

\textbf{Refined scope:} exactly one matching event exists, and the user explicitly
requests changing its time rather than creating, cancelling, or duplicating an event.
&
\textbf{Before refinement:}

\(\widehat{G}(h)=10.4\%\)

\(\widehat{P}_{\mathrm{fail}}(h)=3/16=18.8\%\)

\(\widehat{D}_{\mathrm{reg}}(h)=2.7\%\)

\(\mathrm{Sep}(h)=0.07\)

\textbf{After refinement:}

\(\widehat{G}(h')=7.6\%\)

\(\widehat{P}_{\mathrm{fail}}(h')=0/16=0.0\%\)

\(\widehat{D}_{\mathrm{reg}}(h')=0.8\%\)

Post-commit violation: \(2.0\%\)
&
The initial hypothesis is useful but too broad: target gain is high, but falsification
and regression exceed the default risk thresholds. Because failures are separable,
one precondition-refinement pass preserves \(73.1\%\) of the target gain while
reducing falsification failures to zero. This shows that refinement salvages repairable updates.
\\
\midrule

\textsc{Reject}
&
AgentDojo /
tool-output shortcut
&
\textbf{Scope:} a tool output contains an urgent instruction or an apparent
verification token.

\textbf{Behavior:} follow the tool-suggested action directly without rechecking
the user-level instruction or system constraint.

\textbf{Expected effect:} complete the immediate target task faster by trusting
the tool message.
&
\(\widehat{G}(h)=8.7\%\)

\(\widehat{P}_{\mathrm{fail}}(h)=5/16=31.2\%\)

\(\widehat{D}_{\mathrm{reg}}(h)=4.9\%\)

\(\mathrm{Sep}(h)=0.31\)

Forced-commit violation: \(14.6\%\)
&
Although the hypothesis improves several target-transfer probes, it creates a
high-risk shortcut under decoy and prompt-injection probes. The failures are not
cleanly separable by a narrower precondition. FGSE therefore rejects the update,
preventing a target-only gate from committing an unsafe behavior.
\\

\bottomrule
\end{tabular}%
}
\end{table}

The default FGSE gate commits the round-trip booking rule and accepts the
refined calendar rule after scope narrowing. In the refinement and rejection
cases, falsification and regression evidence govern commitment alongside positive
target-transfer gain. The refinement case demonstrates the practical value of
one-pass scope narrowing:
\(\widehat{G}(h)\) decreases moderately from \(10.4\%\) to \(7.6\%\), but
\(\widehat{P}_{\mathrm{fail}}(h)\) drops from \(18.8\%\) to \(0.0\%\), and
\(\widehat{D}_{\mathrm{reg}}(h)\) drops from \(2.7\%\) to \(0.8\%\). Thus, under the
default setting, FGSE preserves useful self-evolution while blocking the main
failure modes exposed by falsification and regression probes.

\subsection{Mechanistic Analysis of Long-Term Self-Evolution}
\label{app:mechanistic_analysis}

\paragraph{Trajectory-level behavioral changes.}
\label{sec:trajectory_edit_sankey}

Aligned interpretability in autonomous driving ties language to intermediate perception, prediction, and planning outputs~\citep{pmlr-v270-ding25a}.
We compare baseline and FGSE execution trajectories to identify the dominant behavioral edits underlying performance improvements.

Let $\tau^{\mathrm{B}} = (a_1, \dots, a_T)$ and $\tau^{\mathrm{F}} = (a'_1, \dots, a'_{T'})$ denote the baseline and FGSE trajectories for the same task. 
We first align the two traces at the action-type level and group all trajectory edits into four interpretable behavior families:
\emph{tool order}, \emph{memory retrieval}, \emph{constraint checking}, and \emph{final answer formatting}. 
For each family $e \in \mathcal{E}$, we compute its weighted edit mass
\begin{equation}
m_e(\tau^{\mathrm{B}}, \tau^{\mathrm{F}})
=
\sum_{o \in \mathcal{O}_e}
w(o),
\end{equation}
where $\mathcal{O}_e$ is the set of aligned edit operations assigned to family $e$, and $w(o)$ denotes the causal weight of operation $o$ estimated from its contribution to final task improvement.
We then define the dominant edit family as
\begin{equation}
e^\star(\tau^{\mathrm{B}}, \tau^{\mathrm{F}})
=
\arg\max_{e \in \mathcal{E}} m_e(\tau^{\mathrm{B}}, \tau^{\mathrm{F}}).
\end{equation}

\begin{figure*}[t]
    \centering
    \includegraphics[width=0.98\textwidth]{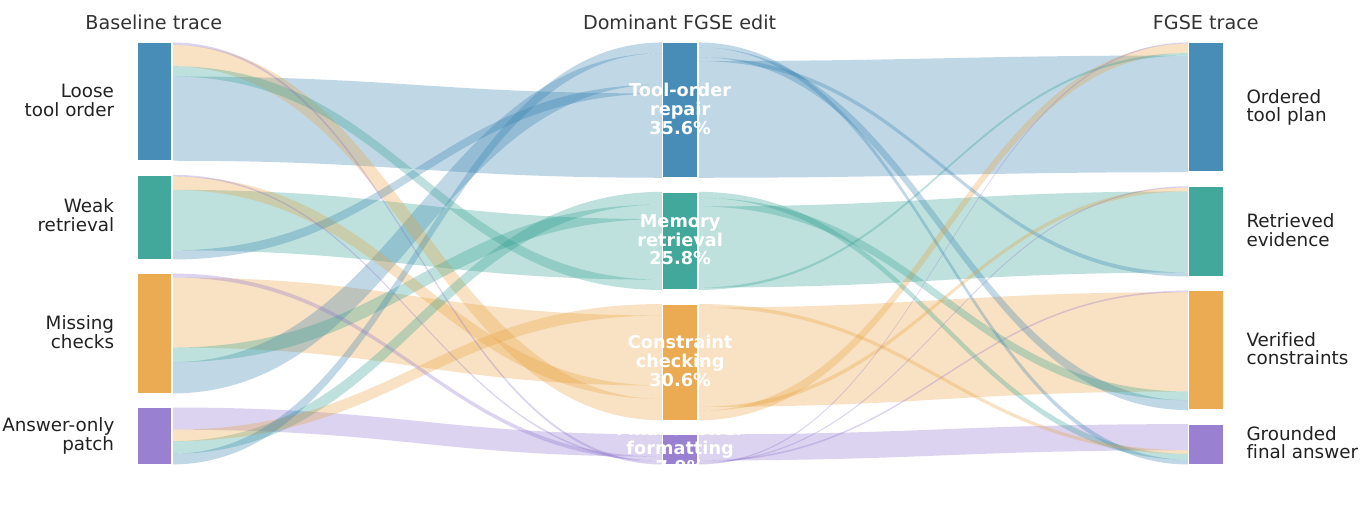}
    \caption{
    \textbf{Trajectory-edit Sankey diagram from baseline traces to FGSE traces.}
    The left column shows the dominant behavioral deficiency in the baseline trace, the middle column shows the dominant edit family introduced by FGSE, and the right column shows the resulting FGSE behavior motif. 
    Band width is proportional to the number of corrected trajectories. 
    The diagram shows that most successful transitions are driven by tool-order repair, memory retrieval, and constraint checking, whereas final-answer formatting accounts for only a small minority of improvements.
    }
    \label{fig:trajectory_edit_sankey}
\end{figure*}

To make the behavioral change interpretable, we further assign each trajectory pair a \emph{baseline behavior motif} and an \emph{FGSE behavior motif}, determined from the dominant failure pattern before the first decisive divergence and the dominant corrected pattern after refinement, respectively. 
Figure~\ref{fig:trajectory_edit_sankey} aggregates all improved trajectory pairs into a three-stage Sankey diagram: baseline motif $\rightarrow$ dominant edit family $\rightarrow$ FGSE motif. 
The width of each band is proportional to the number of corrected trajectories contributing to that transition.

Figure~\ref{fig:trajectory_edit_sankey} shows that FGSE's gains are primarily driven by \emph{substantive behavioral correction}.
The dominant flow mass concentrates in three edit families: tool-order repair, constraint checking, and memory retrieval.
Final-answer formatting occupies only a small portion of the total corrected mass and rarely converts failures to successes.

The largest transition typically starts from baseline traces with loose or misordered tool usage, passes through tool-order repair, and ends in ordered tool plans under FGSE.
A similarly large portion of corrected trajectories flows through constraint checking, showing that many failures arise from missing explicit verification before tool execution or answer commitment despite available knowledge.
Memory retrieval forms the third major band, and its contribution is especially prominent on long-horizon and memory-intensive tasks, where the baseline agent often fails to retrieve previously relevant context or latent facts.

Overall, the Sankey analysis clarifies \emph{where} FGSE changes behavior: it reorganizes tool usage, retrieves missing evidence, and enforces explicit constraint checks.
These findings strengthen the interpretation that FGSE improves the agent's internal decision process.

\subsection{Cost and Scalability}
\label{app:sensitivity_scalability}

\paragraph{Memory-growth controllability.}
\label{sec:memory_growth_pareto}

\begin{figure*}[t]
    \centering
    \includegraphics[width=0.98\textwidth]{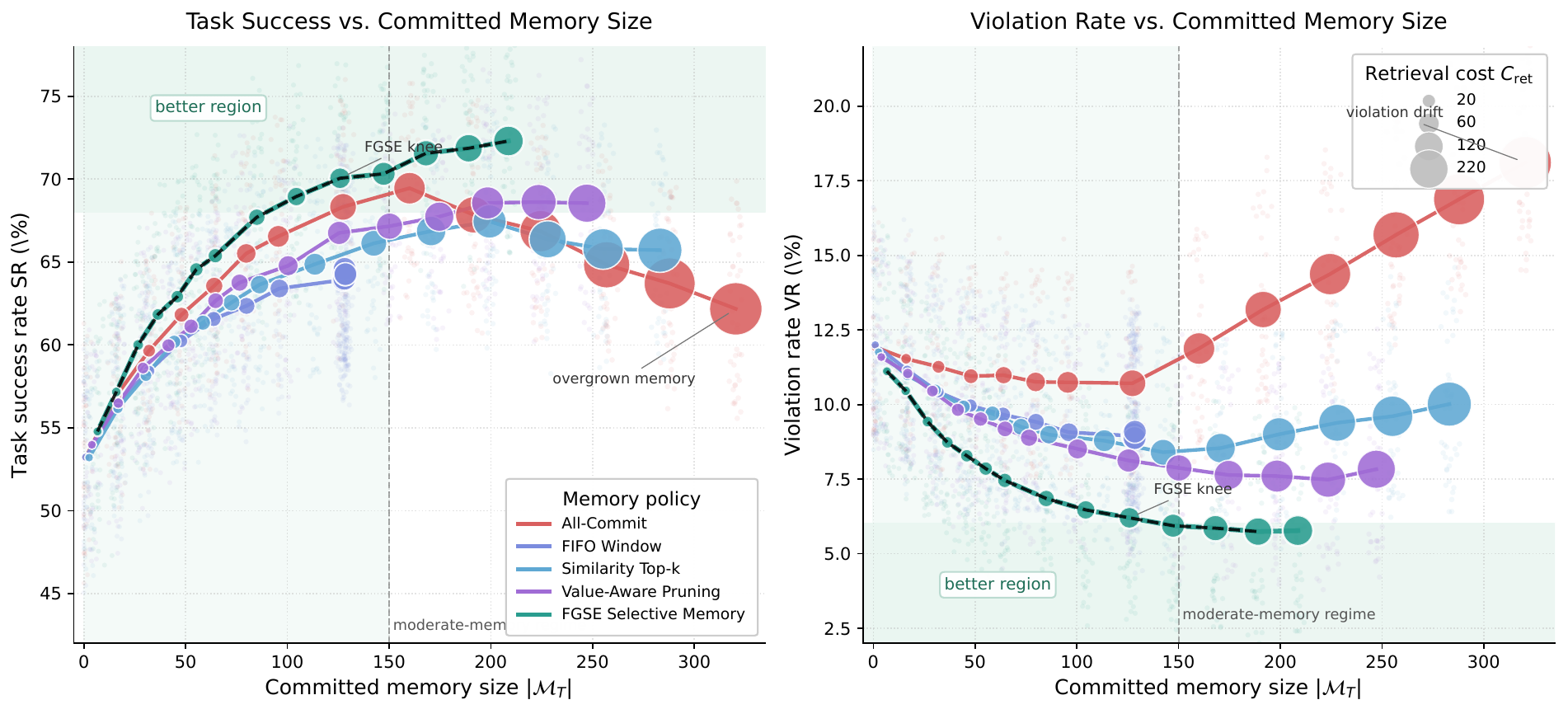}
    \caption{
    \textbf{Memory-growth Pareto analysis.}
    We vary the retention budget and evaluate long-horizon self-evolving agents under different memory policies.
    \textbf{Left:} task success rate versus committed memory size.
    \textbf{Right:} violation rate versus committed memory size.
    Each faint point denotes one domain-seed run, while solid lines connect policy means across budgets.
    Marker area is proportional to retrieval cost $C_{\mathrm{ret}}$.
    A desirable memory policy should remain close to the upper-left region in the left panel (high success with small memory) and the lower-left region in the right panel (low violation with small memory).
    FGSE traces the most favorable frontier: it achieves high task success and low violation with substantially smaller committed memory and lower retrieval cost than naive memory growth.
    }
    \label{fig:memory_growth_pareto}
\end{figure*}

We analyze memory-growth scalability by varying the retention budget and measuring how agent quality changes as the number of committed hypotheses increases.
For each memory policy, we vary the nominal retention budget over
\[
B \in \{0,16,32,48,64,80,96,128,160,192,224,256,288,320\},
\]
and run long-horizon evaluation episodes under the same self-evolution protocol.
For each run, we record the final committed memory size $|\mathcal{M}_T|$, task success rate $\mathrm{SR}$, violation rate $\mathrm{VR}$, and average retrieval cost $C_{\mathrm{ret}}$.
We compare \textsc{All-Commit}, \textsc{FIFO Window}, \textsc{Similarity Top-$k$}, \textsc{Value-Aware Pruning}, and \textsc{FGSE Selective Memory}.
Figure~\ref{fig:memory_growth_pareto} presents the resulting memory-growth Pareto plot.

Figure~\ref{fig:memory_growth_pareto} shows that memory growth is \emph{not} monotonic in utility.
The naive \textsc{All-Commit} strategy initially benefits from additional memory, but after the committed set becomes large, the curve enters an overgrown regime: success saturates and then degrades, while the violation rate rises steadily and retrieval cost increases sharply.
In contrast, bounded heuristics such as \textsc{FIFO Window} and \textsc{Similarity Top-$k$} keep memory smaller, but they also leave substantial performance on the table because they discard useful reusable rules too aggressively.

\textsc{FGSE Selective Memory} lies on the best empirical frontier in both panels.
It reaches its strongest regime at a moderate memory size.
In this diagnostic run, \textsc{FGSE} attains around $70\%\!\sim\!72\%$ task success once the committed memory reaches roughly $120$--$150$ entries, while maintaining a violation rate near $4\%\!\sim\!5\%$.
At a comparable or even lower committed size, competing strategies either achieve lower success or incur substantially higher violation.
Moreover, compared with \textsc{All-Commit}, \textsc{FGSE} achieves a clearly better quality--cost trade-off: it matches or exceeds the best success of naive growth using far fewer retained hypotheses and noticeably smaller retrieval cost.
This indicates that \emph{memory quality and selective retention} are the key bottleneck for gains from persistent memory.

Finally, the curves exhibit a visible saturation effect.
Beyond a moderate memory size, the marginal gain of additional committed hypotheses becomes very small, while retrieval cost continues to increase.
These results support controllable long-term memory growth through commitment filtered by utility, falsification, and regression preservation.

\section{Broader Impacts}\label{app:broader_impacts}
FGSE may improve the reliability of long-horizon language agents by reducing unsafe memory updates, overgeneralized behavioral rules, and regressions on previously solved tasks. 
Potential positive impacts include safer agent adaptation in web, app, and tool-use settings where persistent memory is used. 
However, more reliable self-evolving agents may also increase the capability of automated systems that can perform harmful or unauthorized actions if deployed without access control, monitoring, and domain-specific safety constraints~\citep{yang2026balancing,fan2023towards}.
Financial-agent audits likewise identify source-manipulation risks and motivate model-independent halts and hard loss limits~\citep{wang2026tradingagentsmarketcrashers}. 
FGSE mitigates some of these risks through verifier-grounded falsification, regression testing, audit records, and post-commit monitoring. These mechanisms complement deployment-level safety controls.

\section{LLM Usage}
\label{app:llm_usage}

This paper studies self-evolving LLM agents. In the experiments, frozen LLM
backbones are used as ReAct-style executors that generate interaction
trajectories in web, app, tool-use, and long-term memory environments. FGSE also
uses an LLM-based proposer to convert episode evidence into structured candidate
hypotheses. LLM-generated hypotheses enter verifier-based evaluation. The final
commit--refine--reject decision is determined by target-transfer, falsification,
and regression probes evaluated with executable or benchmark-provided verifiers,
including official success validators, database-state checkers, unit tests,
security validators, and exact-match evaluators.

During the broader research process, LLM-based tools assisted with wording,
grammar, LaTeX formatting, code drafting, script debugging, and organization of
experimental notes. The authors specified, checked, and approved all scientific
claims, methodological choices, mathematical statements, proofs, experimental
settings, reported numbers, figures, tables, and interpretations. They manually
verified all references for existence, relevance, and correctness, and reviewed
and edited LLM-assisted text and code before inclusion in the paper or experiments.

\end{document}